\documentclass[10pt,twocolumn,letterpaper]{article}

\usepackage{iccv}
\usepackage{times}
\usepackage{epsfig}
\usepackage{graphicx}
\usepackage{amsmath}
\usepackage{amssymb}
\usepackage{caption}
\usepackage{tabularx}
\usepackage{algorithm}
\usepackage{algpseudocode}
\newtheorem{problem}{Problem}
\newtheorem{definition}{Definition}
\usepackage{float}
\usepackage{booktabs}
 \usepackage{multirow}
 \usepackage[table]{xcolor}

\usepackage[breaklinks=true,bookmarks=false]{hyperref}

\iccvfinalcopy

\def\iccvPaperID{11665}

\ificcvfinal\fi

\begin{document}
\title{Meta OOD Learning For Continuously Adaptive OOD Detection}
\author{Xinheng Wu$^{\dag}$   \,Jie Lu$^{\dag}$   \,Zhen Fang$^{\dag}$   \,Guangquan Zhang$^{\dag}$\\
$^\dag$Australian Artificial Intelligence Institute, 
University of Technology Sydney\\
{\tt\small Xinheng.Wu@student.uts.edu.au \, \{Jie.Lu, Zhen.Fang, Guangquan.Zhang\}@uts.edu.au} 
\thanks{Corresponding to Jie Lu and Zhen Fang. The work is supported by the
Australian Research Council under Discovery Grant DP$200100700$.}}

\maketitle

\ificcvfinal\thispagestyle{empty}\fi

\begin{abstract}
Out-of-distribution (OOD) detection is crucial to modern deep learning applications by identifying and alerting about the OOD samples that should not be tested or used for making predictions. Current OOD detection methods have made significant progress when in-distribution (ID) and OOD samples are drawn from static distributions. However, this can be unrealistic when applied to real-world systems which often undergo continuous variations and shifts in ID and OOD distributions over time. Therefore, for an effective application in real-world systems, the development of OOD detection methods that can adapt to these dynamic and evolving distributions is essential. In this paper, we propose a novel and more realistic setting called \textit{continuously adaptive out-of-distribution} (CAOOD) detection which targets on developing an OOD detection model that enables dynamic and quick adaptation to a new arriving distribution, with insufficient ID samples during deployment time. To address CAOOD, we develop \textit{meta OOD learning} (MOL) by designing a learning-to-adapt diagram such that a good initialized OOD detection model is learned during the training process. In the testing process, MOL ensures OOD detection performance over shifting distributions by quickly adapting to new distributions with a few adaptations. Extensive experiments on several OOD benchmarks endorse the effectiveness of our method in preserving both ID classification accuracy and OOD detection performance on continuously shifting distributions. 

\end{abstract}

\section{Introduction}

\begin{figure}[t]
\begin{center}
\centerline{\includegraphics[width=\columnwidth]{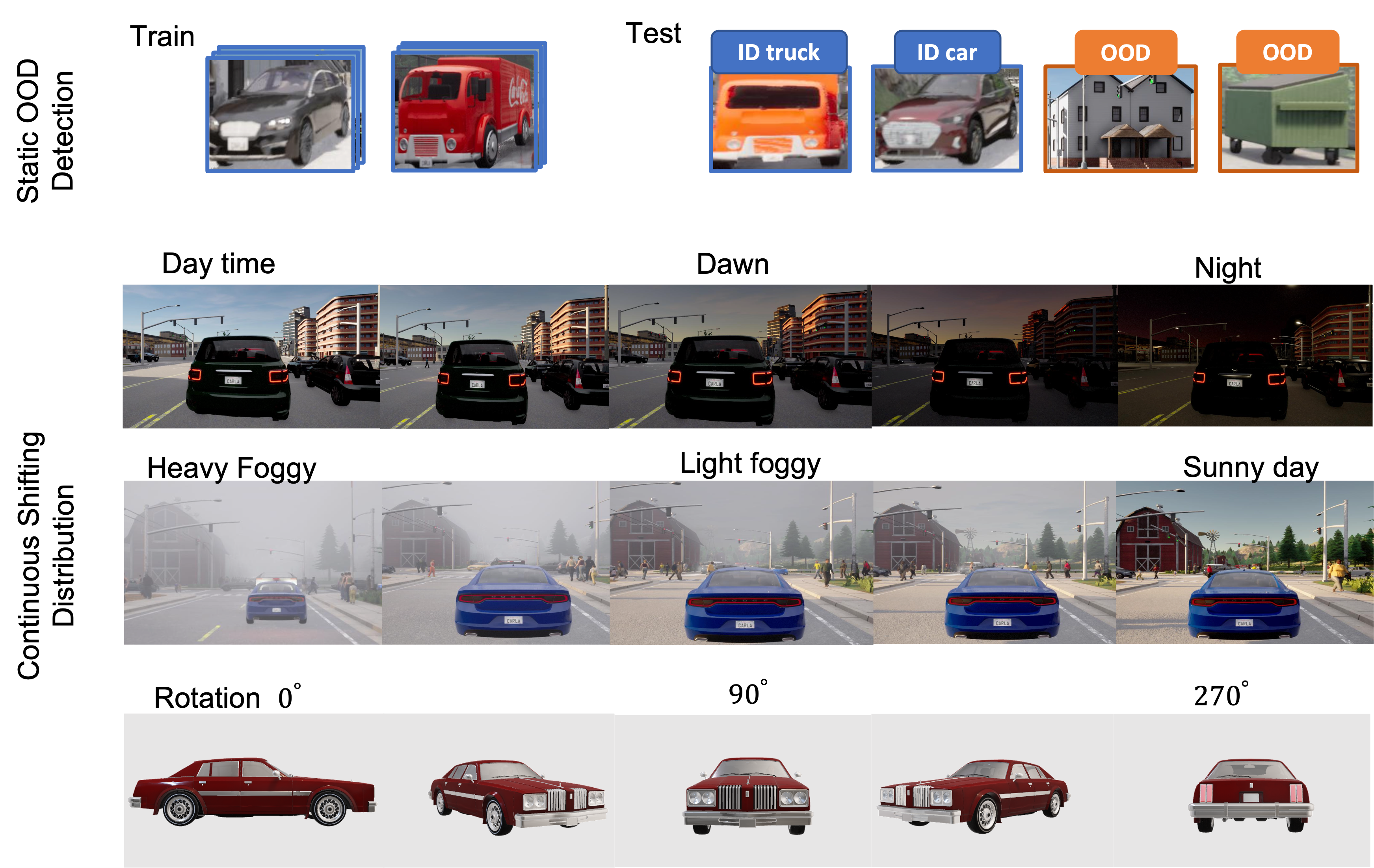}}
\setlength{\belowcaptionskip}{-25pt}
\caption{Difference between existing static OOD detection and a more realistic scenario of Continuously Adaptive OOD Detection (Problem \ref{CAOOD}) when test samples come from continuously shifting distribution, i.e., Definition \ref{CSD}. }\label{F1}
\end{center}
\end{figure}

Out-of-distribution (OOD) detection is vital to modern deep learning (DL) applications in the real world, especially in safety-critical applications such as autonomous vehicle control \cite{huang2020survey}, and medical areas \cite{yang2021generalized}. This is because DL systems may make over-confident and incorrect predictions when encountering the out-of-distribution (OOD) samples, which possess semantic labels that are distinct from those of in-distribution (ID) samples \cite{yang2022openood}. To mitigate this issue, an important problem \textit{out-of-distribution detection} has been proposed and studied extensively \cite{fang2022out,hendrycks2016baseline}. In OOD detection, the classifier is required to perform accurate predictions on ID samples while simultaneously identifying OOD samples. Many representative methods either utilize outputs \cite{hendrycks2016baseline, liu2020energy}, feature representations \cite{lee2018simple, sastry2020detecting}, or gradients \cite{huang2021importance,liang2018enhancing} to enlarge the separations between ID and OOD samples, while others \cite {hendrycks2018deep, ming2022poem} incorporate auxiliary OOD samples to regularize the deep models during training.

Although OOD detection has achieved great progress, existing methods only focus on a simple scenario where ID and OOD samples are assumed to be drawn from \textit{static} ID and OOD distributions. However, many real-world systems are  changing dynamically and thus inherently exhibit continuous distribution shifts over time. Take a self-driving agent equipped with a scene recognition system as an example: in the real world, the surrounding environment (e.g., illumination, weather) may shift continuously on the road, such as from day to night, and from clean to foggy. Treating the arriving training and test samples as from \textit{static} distributions may potentially harm the effectiveness of OOD detection, leading to the misclassification of distribution-shifted ID and OOD samples. Such scenarios highlight the limitations of current OOD detection methods when applied to continuously shifting distributions. Figure \ref{F1} highlights the difference between OOD detection on \textit{static} distribution and when applying to continuously shifting distributions.

To tackle the continuously shifting scenarios in OOD detection, we propose a novel and realistic setting termed \textit{{c}ontinuously {a}daptive {out-of-distribution} (CAOOD) {d}etection}, which targets on developing OOD detection method to 1) quickly adapt to the continuously shifting ID distributions and 2) detect the continuously shifting OOD samples over time. Note that during the deployment time, the ID samples may be insufficient during the adaptation process. Therefore, how to enable \textit{dynamic} and \textit{quick adaptation} with \textit{insufficient} ID samples to achieve good OOD detection performance over time is core challenge of CAOOD detection.

Inspired by domain adaptation \cite{ganin2016domain,9616392_Dong} and meta-learning \cite{finn2017model}, we develop a novel and effective CAOOD detection method called \textit{\underline{m}eta \underline{o}ut-of-distribution \underline{l}earning} (MOL) to address the challenge in CAOOD detection. To dynamically and quickly adapt to the continuously shifting ID samples, we leverage the learning-to-learn paradigm \cite{vilalta2002perspective} of meta-learning which aims to learn an internal representation for quick adaptation to a new task. Specifically, by formulating adaptations to the continuous shifting ID samples as a variety of inner tasks, we design a meta-training procedure for learning to adapt explicitly. Further, to facilitate quick adaptation with insufficient ID samples, we propose to learn a meta-representation during training, which allows us to only update the light-weighted classifier during testing while keeping the meta-representation fixed. Additionally, considering OOD samples are unavailable in training, we propose to generate OOD samples based on the shifting ID feature representations. Lastly, the classifier is trained discriminatively based on both ID and virtual OOD samples.

The contributions are summarized as follows:

\begin{itemize}
\item We propose a more realistic OOD detection setting called CAOOD detection to promote the applications of OOD detection techniques in real-world scenarios.

\item We develop technologies based on meta-learning and domain adaptation to quickly adapt to the continuously shifting ID distributions. With these technologies, MOL is proposed to address CAOOD detection.

\item We conduct extensive experiments comparing MOL with competitive OOD detection methods using various fine-tuning/adaptation strategies on 114 CAOOD detection tasks. Experiments show that MOL achieves the best performance for both ID classification and OOD detection in continuously shifting distributions.
\end{itemize}

\section{Related Work}

\noindent\textbf{Out-of-distribution Detection} methods are mainly post-hoc based adopting different scoring functions based on either logit outputs \cite{hendrycks2016baseline,liang2018enhancing,liu2020energy,sun2021react,lakshminarayanan2017simple,huang2021mos}, feature representations \cite{sastry2020detecting, lee2018simple,lin2021mood,wang2022vim,sun2022out}, or gradients \cite{liang2018enhancing,huang2021importance,igoe2022useful}.  These methods enjoy good applicability when deployed as they can be easily integrated into a pre-trained model without retraining. Other works utilize contrastive learning \cite{tack2020csi,sun2022out,wang2022partial}, specifically designed loss functions \cite{wei2022mitigating, malinin2018predictive, ming2022cider}. However, these methods often require sophisticated score functions but may not necessarily outperform post-hoc methods in general as noted by \cite{yang2022openood}. Another line of OOD detection methods focus on regularizing the training of classifier using either auxiliary OOD datasets or generated fake OOD samples \cite{hendrycks2018deep,lee2018training, liu2020energy}. These works study sampling strategies \cite{ming2022poem,chen2021atom,li2020background}, OOD samples generation approaches \cite{lee2018training,vernekar2019out,duvos}, and uncertainty regularization mechanisms \cite{van2020uncertainty}. In addition to above strategies, other strategies have
also been explored. For example, to overcome the intrinsic inconsistency of prior metrics by aggregating both the known and unknown class performance in a single performance curve, \cite{wang2022openauc} propose a novel OOD detection metric named OpenAUC as the final objective function to learn OOD detector.

Above OOD detection methods neglect inevitable distribution shifts in test data over time in real-world applications. Very few OOD detection works consider such shift: one work \cite{yang2021semantically} used a large unlabeled dataset containing ID samples aiming to encourage ID semantic information modeling while being robust to covariate shift. Another work  \cite{yang2022full} considers explicitly promotes the generalization capacity of the OOD detector when being evaluated on covariate-shifted ID data. Recent work  \cite{ming2022impact} discovers that environmental-related features (e.g., backgrounds) significantly worsen existing OOD detection performance. These early attempts highlight existing OOD detection methods are prone to distribution shifts and consequently inadequate for realistic scenarios, which motivates our research. 

\noindent \textbf{Domain Adaptation (DA)} aims to adapt machine learning models to unseen and different distributions \cite{9695325,yang2021generalized,DBLP:journals/corr/abs-2006-13022,What_Transferred_Dong_CVPR2020}. In complementary to OOD detection, DA improves models' generalization ability to covariate shift when test data share the same label set as training \cite{liu2020open,DBLP:journals/corr/abs-2304-13976}. Most existing DA methods focus on matching discrete source and target domains leveraging domain statistics \cite{pan2010domain}, distance-based loss functions \cite{long2017deep}, and adversarial training \cite{ganin2016domain}. Recently, different methods are developed to perform continuous domain adaptation when the target domain shifts smoothly over time \cite{Bobu2018AdaptingTC, wang2020continuously, wang2022continual}. Continuous domain adaptation is related to our problem but it does not follow an open-world assumption where test data may contain OOD samples that do not belong to the training label set. In this paper, we address a more challenging open-world learning scenario, OOD detection under continuous shifting distributions. 
 
\noindent \textbf{Open World Recognition} aims to incrementally learn new information without forgetting in an open world \cite{joseph2021towards, bendale2015towards,DBLP:conf/icml/FangLLL021}. This problem concentrates on zero-shot learning \cite{rahman2020zero}, mitigating catastrophic forgetting \cite{joseph2021incremental}, and incremental learning \cite{perez2020incremental}, where unknown samples can be progressively labeled as inputs. This is different from our focus on learning an open-world classifier that can quickly adapt to continuously shifted domains online while maintaining both ID and OOD detection performance. Additional literature on the topic of meta-learning is included in the Appendix.

\section{Problem Setup and Notations}
\noindent \textbf{OOD Detection.} Let $\mathcal{X}$ denote the feature space, $\mathcal{Y} =[C]$ \footnote{ We use $[N]$ to represent set $\{1,...,N\}$. Therefore, $[C]=\{1,...,C\}$.} be the label space. We consider the ID distribution $ D_{X_{\rm I}Y_{\rm I}}$ as a joint distribution defined over $\mathcal{X} \times \mathcal{Y}$, where $X_{\rm I}$ and $Y_{\rm I}$ are random variables whose outputs are from spaces $\mathcal{X}$ and $\mathcal{Y}$. Given a set of $n$ samples drawn $\rm i.i.d.$ from the ID distribution called ID data, $S = \{(\mathbf{x}_{j}, y_{j} )\}_{j=1}^{n} \sim D_{X_{\rm I}Y_{\rm I}} $. A classic classification model $\mathbf{f}:\mathcal{X} \to \mathbb{R}^{C} $, is trained on the training set $S$, to predict the label of an input test data \cite{fang2022out}.

During the test time of OOD detection, the test samples contain some unknown OOD samples drawn from an OOD distribution $D_{X_{\rm O}Y_{\rm O}}$, where  $X_{\rm O}$ is a random variable from $\mathcal{X}$, but $Y_{\rm O}$ is a random variable whose outputs do not belong to $\mathcal{Y}$, i.e., $Y_{\rm O} \notin \mathcal{Y}$. The classical OOD detection methods aim to design an effective score function $s(\cdot; \mathbf{f}): \mathcal{X}\rightarrow \mathbb{R}$ \cite{liang2018enhancing, lee2018simple, liu2020energy} and train a corresponding model $\mathbf{f}$ by ID samples $S$ such that the following OOD detector 

\begin{equation}\label{eq1}
G_{\gamma}\left ( \mathbf{x}; {s},\mathbf{f} \right ) = \left\{\begin{matrix}
\hfill  \mathrm{ID} & \mathrm{if} \ s( \mathbf{x} ;\mathbf{f})\geq \gamma  \\ 
\mathrm{OOD}  & \mathrm{if} \ s( \mathbf{x} ;\mathbf{f}) < \gamma 
\end{matrix}\right.
\end{equation}

\noindent where $\gamma$ is a threshold can distinguish ID and OOD samples accurately. In this paper, we select $\gamma$ when 95\% ID data is correctly classified \cite{yang2022openood,DBLP:conf/iclr/WangY0DKLH023,DBLP:conf/nips/WangLZZ0L022}, and use energy score \cite{liu2020energy} as the scoring function to design our OOD detector, i.e., 

\begin{equation}\label{eq2}
s(\mathbf{x};\mathbf{f}) = \log \overset{C}{\sum_{l=1}}\exp(f_l(\mathbf{x})) 
\end{equation}
where $f_l(\mathbf{x})$ is the $l$-th coordinate of $\mathbf{f}(\mathbf{x})$.\\

\noindent \textbf{Continuously Adaptive OOD Detection.} To tackle more realistic scenarios that the ID and OOD samples are from  continuously shifting distributions over a discrete time period $T=\{t_1,t_2,...,t_N\}$, which satisfies that 
$0<t_1<t_2<,...,<t_N$.  We also set $T_k = \{t_1,t_2,...,t_k\}$ and set $T_{K}^- = \{t_{K},t_{K+1},...,t_{N}\}$, where $1\leq k<K\leq N$. It is clear that $T_k \cup T_{K}^- \subset T$ and $T_k \cap T_{K}^- = \emptyset$. Next, the definition of  continuously
shifting ID and OOD distributions is given in Definition \ref{CSD}.

\begin{figure}[t]
\label{caood}
\begin{center}
\centerline{\includegraphics[width=\columnwidth]{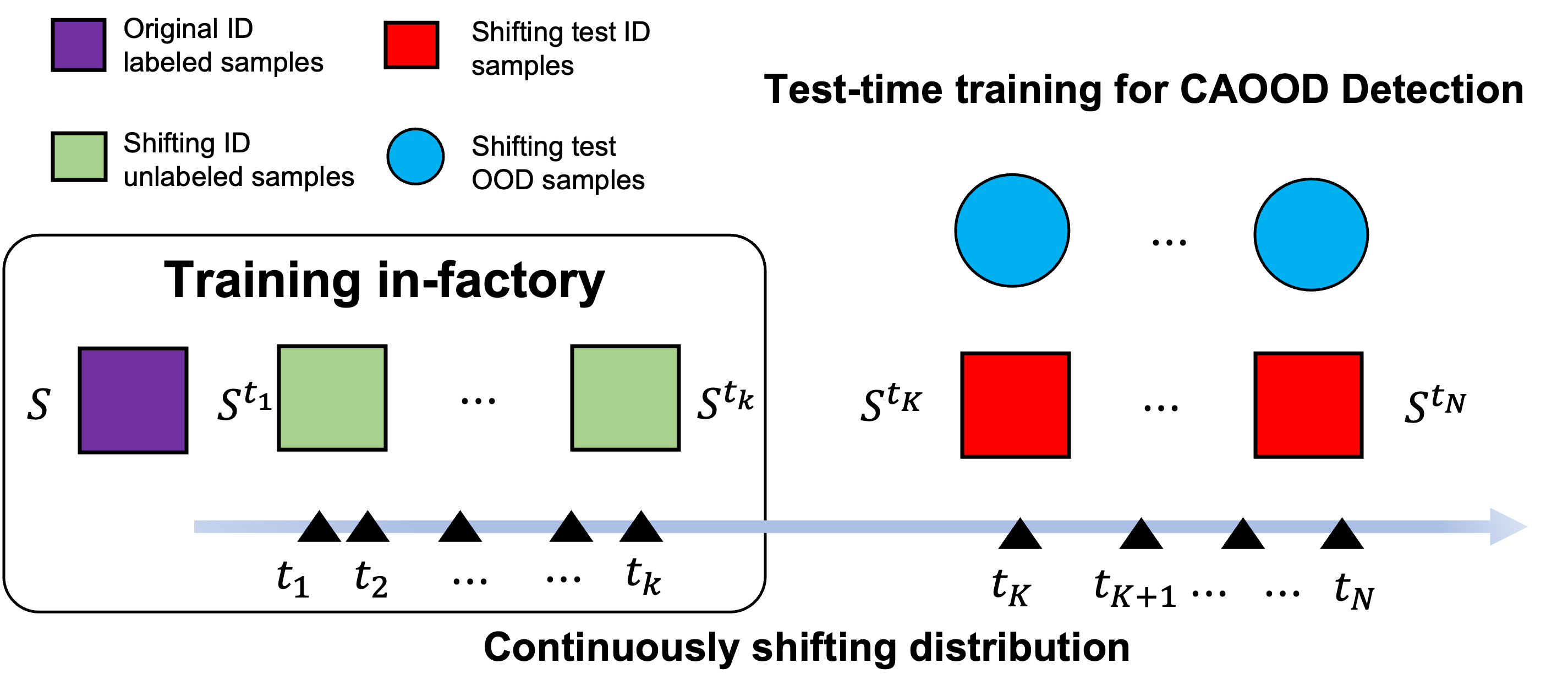}}
\setlength{\belowcaptionskip}{-25pt}
\caption{A heuristic illustration CAOOD problem. The target is to train an OOD detection model $g_0$ using S, $\{S^t\}_{t \in T_k}$ so it can obtain good OOD detection performance on unseen shifting samples $\{S^t\}_{t \in T_K^-}$ by quick adaptation.}\label{F2}
\end{center}
\end{figure}

\begin{definition}[Continuously Shifting Distributions.]\label{CSD}
Let $D^t_{X_{\rm I}Y_{\rm I}}$ and $D^t_{X_{\rm O}Y_{\rm O}}$ are ID and OOD joint distributions at time $t\in T$. We say $D^t_{X_{\rm I}Y_{\rm I}}, D^t_{X_{\rm O}Y_{\rm O}}$ are continuously shifting ID and OOD distributions, if for any $i \in [N-1]$,
\begin{equation*}
\begin{split}
d(D^{t_i}_{X_{\rm I}Y_{\rm I}},D^{t_{i+1}}_{X_{\rm I}Y_{\rm I}})<\epsilon,
\\
d(D^{t_i}_{X_{\rm O}},D^{t_{i+1}}_{X_{\rm O}})<\epsilon,
\end{split}
\end{equation*}
where $d(\cdot,\cdot)$ is a distribution metric and $\epsilon$ is a small value.
\end{definition}

In Definition \ref{CSD}, the small value $\epsilon$ is used to estimate the gradual variations and quantify the continuity of continuously shifting ID and OOD distributions. We note here that distribution shift over time can be significant as it gradually accumulates. Next, we give the definition of continuously adaptive OOD detection.

\begin{problem}[Continuously Adaptive OOD Detection.]\label{CAOOD}
Let $D^t_{X_{\rm I}Y_{\rm I}}$ and $D^t_{X_{\rm O}Y_{\rm O}}$  be the continuously shifting ID and OOD distributions over time period $T$ and let $D_{X_{\rm I}Y_{\rm I}}$ be the original ID distribution. Given sets of samples
\begin{equation*}
\begin{split}
S =  \{(\mathbf{x}_{1}, y_1),..., (\mathbf{x}_{n},y_n)\}    \sim D_{X_{\rm I}Y_{\rm I}},~~i.i.d.
\\
   S^t =  \{\mathbf{x}_{1}^{t} ,..., \mathbf{x}_{n^t}^{t}\}    \sim D_{X_{\rm I}}^t,~~i.i.d.,~\text{for~each}~t\in T
    \end{split}
\end{equation*}
with $n^t \ll n$ if $t\in T^-_K$, the aim of continuously adaptive OOD detection is to train an initial OOD detection model $g_0$ using $S$, $\{S^t\}_{t\in T_k}$, and to quickly update the model $g_{t_i}$ $($at each time $t_i\in T^-_K = \{t_{K},t_{K+1},...,t_{N}\}$$)$ based on the previous model $g_{t_{i-1}}$ $(g_0$ is the initial model, i.e., $g_{t_{K-1}}=g_0$$)$ using $S$, $S^{t_i}$. Such that for any test sample $\mathbf{x} \in \{S^t\}_{t \in T_K^-}$: 1) if $\mathbf{x}$ is from ID distribution $D_{X_{\rm I}}^{t}$, $g_{t_N}$ 
classifies $\mathbf{x}$ into correct ID label; and 2) if $\mathbf{x}$ is from OOD distribution $D_{X_{\rm O}}^{{t}}$, $g_{t_N}$ detects $\mathbf{x}$ as OOD.
\end{problem}

A heuristic illustration of the CAOOD problem is provided in Figure \ref{F2}. The key challenge of Problem \ref{CAOOD} is to quickly adapt to new ID distribution $D_{X_{\rm I}}^{t_i}$ using insufficient ID samples $S^{t_i}$ at time $t_i$. To achieve satisfied performance across time period $T^-_K$, a good initial OOD detection model $g_0$ is indispensable for effective adaptation. In Section \ref{Se::Method}, we illustrate the proposed MOL method and detail how to obtain the initial model $g_0$ during the training procedure. 

\noindent \textbf{Notations.} To facilitate better comprehension, we introduce some necessary notations. To learn a good initial OOD detection model $g_0$, we mainly use meta-learning techniques. Following the meta-training procedure \cite{finn2017model}, we need to sample a support set $\mathbf{S}_{\rm spt} = \{S^{t}_{\rm spt}\}_{t\in T_k'}$ from $\{S^t\}_{t\in T_k}$, where ${T_k'} \subset T_k$ and $S^{t}_{\rm spt}\subset S^t$. We also need to sample a query set $\textbf{S}_{\rm qry} = \{S^{t}_{\rm qry}\}_{t\in T_k'}$ from $\{S^t\}_{t\in T_k}$, where $S^{t}_{\rm qry}\subset S^t$. Further, we set $\mathbf{f}_{\Theta}$ to be the \textit{feature extractor}, set $\mathbf{h}_{\Phi}$ to be the \textit{adapter}, and set $\mathbf{c}_{W}$ to be the \textit{classifier}, where $\Theta, \Phi$ and $W$ are the parameters. Then the classification model $\mathbf{f}$ can be expressed as a function composition: 
\begin{equation*}
\mathbf{f}_{W,\Phi,\Theta} = \mathbf{c}_{W} \circ \mathbf{h}_{\Phi} \circ \mathbf{f}_{\Theta}.
\end{equation*}
 Lastly, the OOD detection model $G_{\gamma}$ can be obtained via the score function $s(\cdot; \mathbf{f}_{W,\Phi,\Theta})$, as shown in Eqs. \ref{eq1} and \ref{eq2}. 

To adapt the shifting distributions, we mainly utilize the \textit{maximum mean discrepancy} (MMD) \cite{gretton2012kernel}, which are introduced in following: given samples $\{\mathbf{x}_i\}_{i=1}^n$ and $\{\mathbf{x}_j'\}_{j=1}^m$,
 \begin{equation*}
 \begin{split}
    & d_k(\{\mathbf{x}_i\}_{i=1}^n, \{\mathbf{x}_j'\}_{j=1}^m)\\ = & \big \| \frac{1}{n} \sum_{i=1}^n \phi_k(\mathbf{x}_i) - \frac{1}{m} \sum_{i=1}^m \phi_k(\mathbf{x}_j) \big \|_k,
     \end{split}
 \end{equation*}
 where $k$ is the kernel and $\phi_k$ is the kernel feature map. To further improve the adaptive performance, in this work, we use the multi-kernel MMD $d_{mk}$, which is shown in \cite{long2015learning}.

\section{Methodology}\label{Se::Method}

\renewcommand{\algorithmicrequire}{\textbf{Input:}}
\renewcommand{\algorithmicensure}{\textbf{Output: }}
\newcommand{\BL}{\par\noindent}
\newcommand{\BLindent}{\par}
\newcommand{\KwTo}{\textbf{to }}
\newfloat{algorithm}{t}{lop}[section]
\floatname{algorithm}{Algorithm}
\setlength{\textfloatsep}{5pt}

It is important to recognize that while the number of test samples $S^t$ ($t\in T_{K}^-$) may be limited during deployment, we often have access to a satisfactory number of labeled training samples $S$ as well as unlabeled samples $\{S^t\}_{t\in T_k}$ that can be used to simulate distribution shifts that may occur during deployment. This motivates us to learn a well-designed initial OOD detection model $g_0$ that utilizes both labeled training samples $S$ and unlabeled samples $\{S^t\}_{t\in T_k}$ to effectively adapt to the shifting test samples $S^t$ ($t\in T_{K}^-$) as defined in Problem \ref{CAOOD}.

To develop an effective initial OOD detection model $g_0$, we consider learning an internal meta-representation that enables efficient adaptation to the new distributions in  Problem \ref{CAOOD}. Additionally, note that OOD samples are unavailable during the training process, we utilize the virtual OOD synthesize \cite{duvos} techniques to generate reliable-virtual OOD samples. The virtual OOD samples emulate real OOD samples and thus enable the classifier to be trained with an additional uncertainty regularization term that incorporates both ID and virtual OOD information. 

An effective strategy of learning $g_0$ is to draw upon the experience of meta-learning \cite{elsken2020meta, vilalta2002perspective}. Originally proposed for addressing few-shot learning \cite{elsken2020meta, jeong2020ood}, meta-learning involves utilizing \textit{fast adaptation} on a small number of samples. By using meta-learning \cite{finn2017model}, we can efficiently adapt the initial model to new distributions encountered during deployment. Specifically, a meta-training process is explicitly designed for learning-to-adapt by viewing the adaptation to each time $t \in {T_{k}'}$ for OOD detection as the inner tasks, thus dynamic adaptations across all time period $T_k'$ can be regarded as one input for the outer-loop training process. Concretely, we can train the classifier by considering both ID adaptation and OOD uncertainty regularization. In the outer loop, we update the model with respect to many inner tasks across all time periods $T_k'$. We detail the inner learning tasks and outer-loop meta-training process in Sections \ref{Se::4.1} and \ref{Se::4.2}. The entire procedures of the proposed MOL method are shown in Algorithms \ref{alg:cap} and \ref{alg:test}.

\subsection{Inner Learning Task for MOL}\label{Se::4.1}
\noindent{\textbf{ID Adaptation.}} In the inner task, we firstly train the adapter $\mathbf{h}_\Phi$ and classifier $\mathbf{c}_W$ to adapt to current ID distribution at time point $t\in T_k'$ using support set $\mathbf{S}_{\rm spt}$. Specifically, given a cross-entropy loss $\ell_{\rm ce}$ and multi-kernel MMD $d_{mk}$, we minimize the following objective.
\begin{equation}\label{eq3}
\begin{split}
  \min_{\Phi,W} \big [ \mathcal{L}_{\rm ce}+\mathcal{L}^t_{\rm d}\big ]=  \min_{\Phi,W} \big [ \frac{1}{n} \sum_{i=1}^{n} \ell_{\rm ce}\left(\mathbf{f}_{\Theta,\Phi,W}\left( \mathbf{x}_i \right) , y_i \right)\\  + d_{mk}^2( \{\mathbf{f}_{\Theta,\Phi,W}(\mathbf{x}_i)\}_{i=1}^n,  \{\mathbf{f}_{\Theta,\Phi,W}(\mathbf{x}_{j,{\rm spt}}^t)\}_{j=1}^{m^t})\big ],
    \end{split}
\end{equation}
where $\{\mathbf{x}_{j,{\rm spt}}^t\}_{j=1}^{m^t}= S_{\rm spt}^t$. Then at each time $t\in T_k'$, we estimate the empirical mean and covariance of training samples $S$ for each class $c \in \mathcal{Y}$ using $\mathbf{h}_{\Phi} \circ \mathbf{f}_{\Theta}$ \cite{lee2018simple},
\begin{equation}\label{eq4}
\begin{aligned}
\centering
 {\hat{\boldsymbol{\mu}}_{c}^t}= \frac{1}{n_{c}} \sum_{i: y_{i}=c} \mathbf{h}_{\Phi} \circ \mathbf{f}_{\Theta} \left( \mathbf{x}_i \right), ~~~
 {\hat{\mathbf{\Sigma}}}^t = \frac{1}{n}\sum_{c=1}^C {\hat{\mathbf{\Sigma}}}_c^t,
\end{aligned}
\end{equation}
where $n_c$ is the number of samples $S$ for class $c$, and
\begin{equation*}
\begin{aligned}
\centering
 {\hat{\mathbf{\Sigma}}}_c^t &= \sum_{i: y_{i}=c}^n \left(\mathbf{h}_{\Phi} \circ \mathbf{f}_{\Theta} \left( \mathbf{x}_i \right) - \hat{\boldsymbol{\mu}}_{c}^t \right) \left( \mathbf{h}_{\Phi} \circ \mathbf{f}_{\Theta} \left( \mathbf{x}_i \right) - \hat{\boldsymbol{\mu}}_{c}^t \right)^\top.
\end{aligned}
\end{equation*}
\noindent Note that the adapter $\mathbf{h}_{\Phi}$ is updated by the training objective in Eq. \eqref{eq3} at each time $t\in T_k'$.

\noindent \textbf{Virtual OOD Generation.} We generate virtual OOD samples based on the updated ID features $\mathbf{h}_{\Phi} \circ \mathbf{f}_{\Theta}$ at each time $t\in T_k'$. Motivated by \cite{duvos,lee2018simple}, we also assume that adapted class-conditional ID distribution $D^t_{\mathbf{h}_{\Phi} \circ \mathbf{f}_{\Theta}(X_{\rm I})|Y_{\rm I}=c}$ is similar to a multivariate Gaussian distribution, i.e.,
\begin{equation*}
    D^t_{\mathbf{h}_{\Phi} \circ \mathbf{f}_{\Theta}(X_{\rm I})|Y_{\rm I}=c} \approx \mathcal{N}\left( \hat{\mathbf{\mu}}_c^t, \hat{\mathbf{\Sigma}^t} \right),
\end{equation*}
which implies that we can sample virtual OOD samples ${Z}_{c}^t$ at the adapted feature space $\mathcal{Z}=\mathbf{h}_{\Phi} \circ \mathbf{f}_{\Theta}(\mathcal{X})\subset \mathbb{R}^{\tilde{d}}$ to each class $c$ in the $\delta$-likelihood region, i.e., ${Z}_{c}^t$ is sampled from
\begin{equation}\label{eq5}
\big \{\mathbf{z}_{c}^{t} \mid \frac{A}{|\hat{\mathbf{\Sigma}}|^{1/2}} \exp \Bigl((\mathbf{z}_{c}^t - \hat{\boldsymbol{\mu}}_{c}^t)\hat{\mathbf{\Sigma}}^{-1}(\mathbf{z}_c^t - \hat{\boldsymbol{\mu}}^{t}_c) \Bigr) < \delta \big \},
\end{equation}
where $A = 1/{(2\pi)}^{{\tilde{d}}/{2}}$ and $\delta$ is a small constant to ensure that the sampled OOD samples ${Z}_{c}^t$ are near the estimated class boundary in the adapted feature space $\mathcal{Z}$.

\noindent \textbf{Inner Uncertainty Regularization.} By the above steps, at each time $t \in T_k'$, we obtained the adapting ID information and the virtual OOD samples at the adapted feature space $\mathcal{Z}$. Now we add an extra regularization term such that the classifier $\mathbf{c}_{W}$ can classify the adapted ID samples and virtual OOD samples in the adapted feature space $\mathcal{Z}$. Specifically, the regularization term \cite{duvos} is shown in following, 
\begin{equation}\label{eq6}
\begin{split}
\mathcal{L}_{\rm ood}^{t} &= \frac{1}{C}\sum_{c=1}^C\mathbb{E}_{\mathbf{z} \sim\mathcal{Z}_{c}^t}\left[ -\log{\frac{1}{1+\exp^{s \left(\mathbf{z};\mathbf{c}_W \right)}}} \right] \\
&+ \mathbb{E}_{\mathbf{x} \sim S^t_{\rm spt}}\left[ -\log{\frac{\exp^{s \left(\mathbf{x};\mathbf{f}_{W,\Phi,\Theta} \right)}}{1+\exp^{ s\left(\mathbf{x};\mathbf{f}_{W,\Phi,\Theta} \right)}}} \right],
\end{split}
\end{equation}
where the score function $s$ is introduced in Eq. \eqref{eq2}. Therefore, we obtain the optimization problem for inner tasks by combining Eq. \eqref{eq3} and \eqref{eq6}, i.e., for a parameter $\lambda>0$,
\begin{equation}
\begin{aligned}
\min_{W, \Phi} \mathcal{L}^t =\min_{W, \Phi} \big [ \mathcal{L}_{\rm ce} + \mathcal{L}^{t}_{\rm d} + \lambda \mathcal{L}_{\rm ood}^{t}\big ].
\label{eq7}
\end{aligned}
\end{equation}

Note that at the earlier training stage, we optimize $\Phi$, $W$ only by using Eq. \eqref{eq3} so that a good estimation of ID distribution can be learned. Specifically, following VOS \cite{duvos},  we maintained a ID class-conditional queue $|Q_y|$ for each class $y \in \mathcal{Y}$ for continuous online estimation of $\hat{\mu}_c^t$ and $\hat{\Sigma}^t$. The uncertainty regularization (Eq. \eqref{eq7}) is introduced in the middle of the training (i.e., at a certain starting epoch $E$).

\subsection{Learning A Meta-representation}\label{Se::4.2}

\noindent \textbf{Outer-loop Training.} In Section \ref{Se::4.1}, the inner training task only involves updating the adapter $\mathbf{h}_{\Phi}$ and classifier $\mathbf{c}_W$, while keeping the feature extractor $\mathbf{f}_{\Theta}$ fixed. In the outer loop, we target on updating $\mathbf{f}_\Theta$ corresponding to fixed $\Phi, W$ given in the inner tasks at each time $t \in T_k'$. Firstly, considering harnessing the knowledge transfer between continuous shifting distributions during time period $T_k'$, we minimize the distribution discrepancy across the time period $T_k'$: let $T_k'= \{t_1',t_2',...,t_l'\}$ with $t_1'<t_2'<,...,<t_l'$, then
\begin{equation}
\begin{aligned}
 \mathcal{L}_{\rm qry} = \max_{i\in [l-1]} d_{mk}^2 \big (\mathbf{f}_{\Theta,\Phi,W}(S_{{\rm qry}}^{t_{i+1}'}),  \mathbf{f}_{\Theta,\Phi,W}(S_{{\rm qry}}^{t_i'})\big),
\end{aligned}
\end{equation}
where $\mathbf{f}_{\Theta,\Phi,W}(S_{{\rm qry}}^{t_{i}'})= \{\mathbf{f}_{\Theta,\Phi,W}(\mathbf{x}_{j,{\rm qry}}^{t_i'})\}_{j=1}^{m^{t_i'}}$, here we set $S_{{\rm qry}}^{t_{i}'}= \{\mathbf{x}_{j,{\rm qry}}^{t_i'}\}_{j=1}^{m^{t_i'}}$.  Intuitively, $\mathcal{L}_{\rm qry}$ estimates the discrepancies of continuously shifting distributions across the time period $T_k'$. Then, the optimization issue in outer-loop training can be represented as follows:
\begin{equation}\label{eq9}
\begin{split}
\min_{\Theta} \mathcal{L}_{\rm meta} = \min_{\Theta} & \big [\mathcal{L}_{\rm ce} + \mathcal{L}_{\rm qry}\\& +  \frac{1}{|T_k'|} \sum_{t\in T_k'} \big ( \mathcal{L}^t_{\rm d}+\lambda \mathcal{L}^{t}_{\rm ood} \big) \big ],
\end{split}
\end{equation}
where $\mathcal{L}_{\rm d}^t$ and $\mathcal{L}^{t}_{\rm ood}$ are computed by using the samples $S_{\rm qry}^t$ in the query set $\mathbf{S}_{\rm qry}$.

In Eq. \eqref{eq9}, the inner tasks (corresponding to $\mathcal{L}^t_{\rm d}+\lambda \mathcal{L}^{t}_{\rm ood}$) across all time period $T_k'$ are fed as one input (corresponding to $ \sum_{t\in T_k'} \big ( \mathcal{L}^t_{\rm d}+\lambda \mathcal{L}^{t}_{\rm ood} \big )/{|T_k'|}$) in the outer-loop training. Thus, we are able to learn a good initialization of meta-representation $\Theta$ such that: when encountering a new distribution during testing, a few updating steps of $\Phi, W$ result in a good performance.\\

\begin{algorithm}[t]
\small
\caption{. MOL in Training Process}\label{alg:cap}
\begin{algorithmic}
\Require ID training samples $S$ and $\{S^t\}_{t\in T_k}$; learning rates $\alpha$, $\beta$; \BL randomly initialized model $\mathbf{f}_{W,\Phi,\Theta} = \mathbf{c}_{W} \circ \mathbf{h}_{\Phi} \circ \mathbf{f}_{\Theta}$.
\While{$ not \ done$}
    \State Randomly initialized adapter $\mathbf{h}_{\Phi}$ and classifier $\mathbf{c}_{W}$;
    \State Randomly sample $T_k'= \{t_1',t_2',...,t_l'\}$ from $T_k$  and sample $~~~~~~$ support and query sets $\mathbf{S}_{\rm spt}$ and $\mathbf{S}_{\rm qry}$ from $\{S^t\}_{t\in T_k'}$;
    \For{$i = 1$ \KwTo $l$}
        \State {Estimate} $\hat{\boldsymbol{\mu}}_{c}^{t_i}$ and ${\hat{\mathbf{\Sigma}}}^{t_i}$ by using $S$ and Eq. \eqref{eq4};
        \State {Sample} virtual OOD samples $Z_c^{t_i}$ by Eq. \eqref{eq5};
        \State \textbf{Compute} $\mathcal{L}^{t_i}$ using $S$, $S^{t_i}_{\rm spt}$ and $Z_c^{t_i}$ by Eq. \eqref{eq7};
        \State \textbf{Update} parameters: $(\Phi, W)_{t_{i+1}} =(\Phi, W)_{t_i} - \alpha \triangledown \mathcal{L}^{t_i}$;
    \EndFor    
    \State \textbf{Compute} $\mathcal{L}_{\rm meta}$ using $S$ and $\mathbf{S}_{\rm qry}$ by Eq. \eqref{eq9};
    \State \textbf{Update} parameters: $\Theta = \Theta - \beta \triangledown \mathcal{L}_{\rm meta}$;
\EndWhile \BL
\algorithmicensure initial model $G_{\gamma}\left ( \mathbf{x}; {s},\mathbf{f}_{\Theta,\Phi,W} \right )$  by Eq. \eqref{eq1} \\ $~~~~~~~~~~~~~~$ \text{meta-representation } $\mathbf{f}_{\Theta}$.
\end{algorithmic}
\end{algorithm}

\begin{algorithm}[t]
\raggedleft
\small
\caption{. MOL in Testing Process}\label{alg:test}
\begin{algorithmic}
\Require ID training samples $S$ and $\{S^t\}_{t\in T_K^-}$; learning rates $\alpha$, $\beta$; \BL meta-representation $\mathbf{f}_{\Theta}$; score function introduced in Eq. \eqref{eq2}.
\While{$ not \ done$}
    \State Randomly initialized adapter $\mathbf{h}_{\Phi}$ and classifier $\mathbf{c}_{W}$;
    \For{$i = K$ \KwTo $N$}
        \State Estimate $\hat{\boldsymbol{\mu}}_{c}^{t_i}$ and ${\hat{\mathbf{\Sigma}}}^{t_i}$ by using $S$ and Eq. \eqref{eq4};
        \State Sample virtual OOD samples $Z_c^{t_i}$ by Eq. \eqref{eq5};
        \State \textbf{Compute} $\mathcal{L}^{t_i}$ using $S$, $S^{t_i}$ and $Z_c^{t_i}$ by Eq. \eqref{eq7};
        \State \textbf{Update} parameters: $(\Phi, W)_{t_{i+1}} =(\Phi, W)_{t_i} - \alpha \triangledown \mathcal{L}^{t_i}$;
    \EndFor    
     \State \textbf{Output:} model $G_{\gamma}\left ( \mathbf{x}; {s},\mathbf{f}_{\Theta,\Phi,W} \right )$ by Eq. \eqref{eq1}.
\EndWhile \BL
\end{algorithmic}
\end{algorithm}

\noindent \textbf{Meta-testing.} In the testing process, when exposed to previously unseen shifting distributions $D^{t}_{X_{\rm I}Y_{\rm I}}$ (where $t\in T_{K}^-$), we only need to fine-tune the adapter $\mathbf{h}_{\Phi}$ and classifier $\mathbf{c}_W$ on the new test samples $S^t$ (where $t\in T_{K}^-$), while keeping the meta-representation $\mathbf{f}_{\Theta}$ fixed. This strategy enables fast adaptation by updating the lightweight classifier and adapter, which is useful for real-world online applications like self-driving agents. Furthermore, by incorporating the score function in Eq. \eqref{eq2}, we can obtain an OOD detection model $g_t$, while updating $\mathbf{h}_{\Phi}$ and $\mathbf{c}_W$ at each time $t\in T_{K}^-$ during the testing process.

\section{Experiments}

\subsection{Adaptive OOD Benchmark Construction}

For CAOOD evaluation, we construct 3 benchmarks (e.g., using Rotation MNIST, Cifar Corruption datasets as ID datasets) derived from commonly used static OOD benchmarks (e.g., MNIST, Cifar). Rotation MNIST and Corruption datasets are frequently used in continuous domain adaptation \cite{wang2020continuously,wang2022continual}. In practical applications, rotation can replicate shifts arising from camera positions, while corruptions mimic various weather conditions (e.g., fog, snow) and movements (e.g., motion). In CAOOD, we evaluate both ID accuracy and OOD detection effectiveness. Dataset details are provided in the Appendix.

\textbf{Rotation MNIST \cite{deng2012mnist}.} This dataset has MNIST digits with various rotations from  $\left[0, 180^{\circ}\right]$, and for each rotation, there are 60,000 training images. In our benchmark, we use Rotation MNIST (R-MNIST) as the ID training dataset and evaluate the OOD detection performance on Rotation NOTMNIST (R-NOTMNIST) and Rotation Cifar10bw (R-Cifar10bw). There are no semantics overlaps between the ID and OOD datasets. 

In our proposed MOL protocol, we consider images with rotation $0^{\circ}$ as the labeled original training samples $S$, and images from rotation $\left(0, 180^{\circ}\right]$ as continuously shifting distributions over the whole time period $\left(0, T \right]$, i.e., $\{S^t\}_{t\in T}$. In meta-training, we use images from rotation $\left(0 - 60^{\circ}\right]$ as $\{S^t\}_{t\in T_k}$, and randomly sample $\mathbf{S}_{\rm spt}$, $\mathbf{S}_{\rm qry}$ of length 10. During meta-testing, we update our model online for rotation $\{S_{120^{\circ}}, S_{126^{\circ}}, i...S_{174^{\circ}}\}$, i.e., $\{S^t\}_{T_K^-}$ and for each rotation, we only use 100 training samples for adaptation. 

\textbf{Cifar10C.} This dataset consists of 15 types of corruptions (e.g., fog, brightness, motion, noise) with each demonstrating 5 levels of severity. We use Cifar10C as the ID data and test on two near OOD datasets applying the same shifting distribution: TinyimageNetC, and Cifar100C. Following the OOD benchmark literature \cite{yang2022openood}, we create TinyimageNetC from a subset of the Tinyimagenet \cite{torralba200880} where 1207 images overlap semantic labels with Cifar10 are removed. 

In our protocol, we take clean images from Cifar10 as original training samples $S$, and consider images with various corruptions from Cifar10C as samples come from shifting distributions. Specifically, we design the continuous shifting distributions by gradually changing the severity across all corruption types, for example: 
$\scriptsize \underset{\text{t-1 and before}}{\underbrace{{\cdots},\mathrm{C}_{t-1}^{5}}} {\rightarrow}\  \underset{\text{corruption type t, changing gradually}}{\underbrace{\mathrm{C}_{t}^{1},\mathrm{C}_{t}^{2},\mathrm{C}_{t}^{3},\mathrm{C}_{t}^{4},\mathrm{C}_{t}^{5},\mathrm{C}_{t}^{4},\mathrm{C}_{t}^{3},\mathrm{C}_{t}^{2},\mathrm{C}_{t}^{1} }} {\rightarrow}\  \underset{\text{t+1 and on}}{\underbrace{\mathrm{C}_{t+1}^{1},\mathrm{C}_{t+1}^{2},{\cdots}}}$

\noindent In meta-training, we randomly sample $\mathbf{S}_{\rm spt}$, $\mathbf{S}_{\rm qry}$ of length 10 (i.e., $|T_{k}^{'}|=10$), from the first 7 corruptions of 63 continuously shifting distributions, i.e., $\{S^t\}_{t\in T_k}, |T_k|=63 $. In meta-testing, we adapt our model to a trajectory of length 10 that is randomly sampled from the last 7 corruptions and evaluate on unseen test images from the last 7 corruptions (i.e., frost, fog, brightness, contrast, elastic transform, pixelate, jpeg compression). Similarly, we only had access to 100 training samples in meta-testing.

\noindent \textbf{Cifar100C \cite{hendrycksbenchmarking}.} This dataset is a variant of Cifar10C by applying the same corruptions to Cifar100. We used Cifar100C as the ID dataset and two near OOD datasets with the same continuous shifting distributions: Cifar10C and TinyimageNetC. We re-create TinyimageNetC after 2505 images sharing the same classes with Cifar100 have been removed \cite{yang2022openood}. For the MOL protocol, we applied the same training and testing fashion as used in Cifar10C. 

\noindent \textbf{Standard OOD Datasets.} In addition, we evaluate our method's OOD detection performance on commonly used OOD datasets from a static distribution, including Cifar10 and Cifar100 \cite{krizhevsky2009learning}, TinyImageNet \cite{torralba200880}, Textures \cite{kylberg2011kylberg}, LSUN-Resize, LSUN-Crop \cite{yu2015lsun}, and iSUN \cite{xiao2010sun}. Table \ref{table1} summarizes evaluated OOD detection benchmarks.

\begin{table}
\centering
\caption{OOD benchmarks used in our evaluation, including CAOOD datasets, near OOD, and far OOD datasets.}
\resizebox{\columnwidth}{!}{\begin{tabular}{l l l l}
 \toprule
 \label{table1}
 ID Dataset& CAOOD & Near OOD & Far OOD  \\
 \midrule
\multirow{2}{*}{R-MNIST} & R-NOTMNIST & NOTMNIST  & Cifar10bw \\
& R-Cifar10bw  & & \\
  \hline
 \multirow{2}{*}{Cifar10C}  & Cifar100C & Cifar100 & Textures   \\
  & TinyimagenetC & TinyimageNet  & LSUN \\
  & & & iSUN  \\
  \hline
 \multirow{2}{*}{Cifar100C} & Cifar10C  & Cifar10 & Textures  \\
 & TinyimagenetC  & TinyimageNet & LSUN  \\
  &  & & iSUN \\
\bottomrule
\end{tabular}}
\end{table}

\begin{table*}[t]
\centering
\caption{Main Results on Rotation MNIST. $\uparrow$ (or $\downarrow$) indicates greater (or smaller) values are preferred. For each comparable method, we report results on Direct Test / Simple Adaptive  / Domain Adaptation. Only Direct Test results are reported for Gram and KNN, only Direct Test and Simple Adaptive results are reported for VOS and LogitNorm. The bold and * represent the best and second best performance and the shadow part marks our method. }\label{table2}
\resizebox{\textwidth}{!}{
\begin{tabular}{@{} c  c c c c c c c @{}}
\toprule
 \multirow{2}{*}{Method}  & \multicolumn{2}{c}{Rotation NOTMNIST} &  \multicolumn{2}{c}{Rotation Cifar10bw} & \multicolumn{2}{c}{Average} & \multirow{2}{*}{ID Accuracy} \\\cmidrule(l){2-7}
   & AUROC $\uparrow$ & FPR95 $\downarrow$ & AUROC $\uparrow$ & FPR95 $\downarrow$ & AUROC $\uparrow$ & FPR95 $\downarrow$ \\
\midrule
  MSP  & 76.9 / 75.9  /62.3  &	71.9 / 70.4 / 87.6 &	90.1 / 89.4 / 68.8 &	55.6 / 48.5 / 83.6 & 83.5 / 82.6 / 65.6 & 64.8 / 59.5 / 85.6 &  25.6 / 28.3 / 27.2 \\
 ODIN & 63.6 /76.7 / 30.8 & 85.2 / 72.6/ 56.7 & 87.3 / 88.1 / 67.6 &58.2 / 50.9 / 83.9 & 75.5 / 82.4 / 49.2 & 71.7 / 61.8 / 70.3 & 25.6 / 28.3 / 27.2\\
 Mahalanobis & 61.2 / 75.4 / 32.9 & 86.1 / 75.1/ 56.9 & 89.3 / 90.1 / 66.7 & 58.1 / 49.5 / 80.8 & 75.3 / 82.8 / 49.8 & 72.1 / 62.3 / 68.9 & 25.6 / 28.4 / 27.7\\
 Energy & 92.6 / 91.7 / 72.7	& 36.1 / 39.7 / 76.1	& 97.5 / 95.9 / 69.1 & 12.6 / 19.9 / 85.2 & 95.1 / 93.8 / 70.9 & 24.4 / 29.8 / 80.7 & 25.6 / 28.3 / 27.2  \\
 Gram  & 96.1$^{*}$ & \textbf{10.6}& 98.9$^{*}$ & 4.6$^{*}$ & 97.5$^{*}$ & \textbf{7.6} & 25.6 \\
 VOS  & 86.7 / 93.5  & 57.6 / 32.3  & 92.7 / 96.7  & 31.0 / 16.4 & 89.7 / 95.1 & 44.3 / 24.4 & 27.7 / 30.9 \\

 LogitNorm & 84.0 / 94.0  & 46.2 / 29.7  & 97.7 / \textbf{99.0}  & 10.5 / \textbf{4.5}  & 90.0 / 96.5  & 28.3 / 17.1  & 24.7 / 31.5$^{*}$  \\
 KNN & 95.7  & 16.5  & 91.4  & 25.0  & 93.5  & 20.8 & 24.9 \\
\rowcolor[HTML]{EFEFEF} 
MOL &\textbf{96.5} & 13.9$^{*}$ & 98.9$^{*}$ & 9.2  & \textbf{97.7} & 11.5$^{*}$ & \textbf{35.6} \\
\bottomrule
\end{tabular}}
\end{table*}

\begin{table*}[t]
\centering
\caption{Main Results on Cifar10 corruption. $\uparrow$ (or $\downarrow$) indicates greater (or smaller) values are preferred. For each comparable method, we report results on Direct Test / Simple Adaptive / Domain Adaptation. Only Direct Test results are reported for Gram, Gradnorm, and KNN, only Direct Test and Simple Adaptive results are reported for VOS and LogitNorm. The bold and * represent the best performance and the shadow part marks our method.}\label{table3}

\resizebox{\textwidth}{!}{
\begin{tabular}{@{} c  c c c c c c c @{}}
\toprule
 \multirow{2}{*}{Method}  & \multicolumn{2}{c}{Cifar100C} &  \multicolumn{2}{c}{TinyImagenetC} & \multicolumn{2}{c}{Average} & \multirow{2}{*}{ID Accuracy} \\\cmidrule(l){2-7}
   & \multicolumn{1}{c}{AUROC $\uparrow$} & \multicolumn{1}{c}{FPR95 $\downarrow$}  & AUROC $\uparrow$ & FPR95 $\downarrow$ & AUROC $\uparrow$ & FPR95 $\downarrow$ & \\
\midrule
 MSP & 53.2 / 62.2 / 54.8 & 94.3 / 91.6 / 92.9 & 59.7 / 64.0 / 52.9 & 92.5 / 91.2 / 94.9 & 56.5 / 63.1 / 53.8 & 93.4 / 91.4 / 93.9 & 38.5 / 56.2 / 35.6 \\
 ODIN &  46.1 / 61.3 / 55.2 & 96.5 / 91.9 / 93.4 & 40.3 / 62.9 /52.0  & 99.1 / 91.5 / 94.9 & 43.2 / 62.1 / 53.6  & 97.8 / 91.7 / 94.2 & 38.1 / 55.6 / 35.2 \\
 Mahalanobis & 46.8 / 63.6 / 55.1 & 96.7 / 91.3 / 93.4 & 47.0 / 66.0 / 53.6 & 98.1 /  90.6 / 93.9 & 46.9 / 64.8 / 54.4 & 97.4 / 90.9 / 93.7 & 38.5 / 56.2 / 35.6\\
 Energy  & 52.7 / 60.8 / 56.5 & 94.4 / 91.4 / 93.7 & 61.5 / 65.0 / 60.8 & 91.6 / 89.0 / 94.6 & 57.1 / 62.9 / 58.7 & 93.0 / 90.2 / 94.2 & 38.5 / 52.9 / 35.6	\\
 Gram & 55.2 & 89.7 &	66.5 & \textbf{82.5} & 60.9 & 86.1$^{*}$ & 38.1\\
 Gradnorm  & 29.5 &	97.2 & 41.3 & 96.5 & 35.4 & 96.9 & 38.5\\
 VOS & 52.6 / 53.6 & 95.7 / 96.5  & 57.3 / 59.9  & 97.2 / 97.1  & 54.9 / 56.8  & 96.5 / 96.8 & 25.9 / 31.7 \\
 
 LogitNorm & 56.4 / 65.3$^{*}$  & 94.5 / 88.9$^{*}$  & 61.9 / 67.6$^{*}$  & 92.6 / 85.9  & 59.1 / 66.4$^{*}$  & 93.6 / 87.4 & 43.5 / 59.1$^{*}$  \\
 KNN& 54.1 & 94.4  & 57.7 & 93.5  & 55.9  & 94.0  & 44.2 \\
\rowcolor[HTML]{EFEFEF} 
MOL & \textbf{69.7} & \textbf{86.3} & \textbf{71.4} & 85.6$^{*}$ & \textbf{70.6} & \textbf{85.9} &  \textbf{64.1}	 \\
\bottomrule
\end{tabular}}
\end{table*}

\begin{table*}[ht]
\centering
\caption{Main Results on Cifar100 corruption. $\uparrow$ (or $\downarrow$) indicates greater (or smaller) values are preferred. For each comparable method, we report results on Direct Test / Simple Adaptive / Domain Adaptation (DA). Only Direct Test results are reported for Gram, Gradnorm, and KNN, only Direct Test and Simple Adaptive results are reported for VOS and LogitNorm. The bold and * represent the best performance and the shadow part marks our method.}\label{table4}

\resizebox{\textwidth}{!}{
\begin{tabular}{@{} c  c c c c c c c @{}}
\toprule
 \multirow{2}{*}{Method}  & \multicolumn{2}{c}{TinyImagenetC} &  \multicolumn{2}{c}{Cifar10C} & \multicolumn{2}{c}{Average} & \multirow{2}{*}{
ID Accuracy} \\\cmidrule(l){2-7}
   & \multicolumn{1}{c}{AUROC $\uparrow$} & \multicolumn{1}{c}{FPR95 $\downarrow$}  & AUROC $\uparrow$ & FPR95 $\downarrow$ & AUROC $\uparrow$ & FPR95 $\downarrow$ & \\
\midrule
 MSP & 50.9 / 61.3 / 58.1 & 94.8 / 91.3 / 92.5 & 58.2 / \textbf{64.0} / 61.2 & 92.9 / 91.2 / 91.3 & 54.6 / 62.7$^{*}$ / 59.7 & 93.4 / 91.3 / 91.9 & 27.8 / 41.4 / 37.7 \\
 ODIN &  49.1 / 59.7 / 55.5  & 95.5 / 91.4 / 93.6 & 41.8 / 63.5 / 60.2  & 98.0 / 90.8$^{*}$ / 91.6 & 45.5 / 61.6 / 57.9  & 96.8 / 91.1$^{*}$ / 92.2   & 27.8 / 41.4 / 37.7 \\
 Mahalanobis & 48.5  & 97.1  & 40.4  & 99.0  & 44.5 & 98.1 & 27.8 \\
 Energy  & 50.7 / 58.0/ 61.6$^{*}$ & 94.9 / 92.1 / 91.0$^{*}$ & 58.7 / 63.6$^{*}$ / 61.0 & 92.9 / 91.4 / 92.6 & 54.7 / 60.8 / 61.3 & 93.9 / 91.8 / 91.8 & 27.8 / 41.4 / 37.7	\\
 Gram &  47.1 & 95.9 & 49.6 & 95.6 & 48.4 & 95.8 & 27.8 \\
 VOS & 49.9 / 52.1  & 96.0 / 96.3  &  52.6 / 54.8  & 97.7 / 98.0  & 51.3 / 53.5 & 96.9 / 97.2  & 26.1 / 43.1$^{*}$  \\

 LogitNorm & 53.0 / 58.3  & 94.1 / 92.2  & 58.8 / 63.1  & 93.2 / 91.9  & 55.9 / 60.7  & 93.7 / 92.0  & 23.4 / 38.5  \\
 KNN & 51.7  & 95.1  & 50.0  & 94.9  & 50.9  & 95.0  & 22.9 \\
\rowcolor[HTML]{EFEFEF} 
MOL & \textbf{69.4}& \textbf{88.7} & 63.1 & \textbf{89.0} & \textbf{66.3} & \textbf{88.9} & \textbf{57.4}	 \\
\bottomrule
\end{tabular}}
\end{table*}

\begin{figure*}[t]
\begin{center}
\centerline{\includegraphics[width=\textwidth]{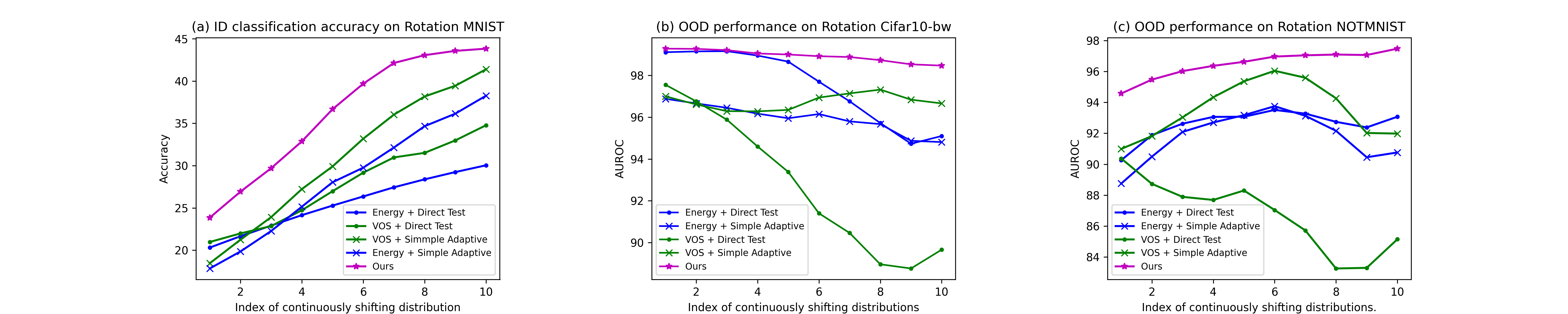}}
\setlength{\belowcaptionskip}{-20pt}
\caption{The ID classification (a) and OOD detection performance (b-c) on R-MNIST benchmark over continuously shifting distributions. Our method achieved the best performance on all unseen distributions, i.e., the pink line.}\label{F3}
\end{center}
\end{figure*}

\subsection{Baselines and Metrics}
We compare our method with comprehensive OOD detection baselines including Maximum Softmax Probability \cite{hendrycks2016baseline}, ODIN \cite{liang2018enhancing}, Mahalanobis distance \cite{lee2018simple}, energy score \cite{liu2020energy}, Gram Metrics \cite{sastry2020detecting}, Gradnorm \cite{huang2021importance}, and the recent VOS \cite{duvos}, KNN \cite{sun2022out} and LogitNorm \cite{wei2022mitigating}. To apply the above static OOD detection baselines in CAOOD, we use three different training/adapting schemes as follows. 

\noindent \textbf{Direct Test.} The model is trained on the labeled original training samples $S$ and then directly test on continuously shifting $\{S^t\}_{t \in T_K^-}$ for OOD detection with no adaptation. 

\noindent \textbf{Simple Adaptive.} During training, the model is trained on the labeled original training samples $S$. During testing, we finetune the classifier of the model to adapt to continuously shifting $\{S^t\}_{t \in T_K^-}$ dynamically and test their performance. Specifically, the classifier is updated using our proposed meta-testing strategy without virtual OOD generation and uncertainty regularization (i.e., by using Eq. \ref{eq3}).

\noindent \textbf{Domain Adaptation.} The model is trained using classic domain adaptation techniques DAN \cite{long2015learning}. Specifically, in the training process, we train the model offline by viewing $S$ as the source domain and all shifting training samples $\{S^t\}_{t \in T_k}$ available as the target domain. During testing, we finetune the classifier of the model to adapt to continuously shifting $\{S^t\}_{t \in T_K^-}$ dynamically and test their performance (Eq. \eqref{eq3}). By doing so the Domain Adaptation strategy violates the CAOOD setting during training.

Note that for Gram Metrics, Gradnorm, and KNN, only the Direct Test results are reported because their scoring functions are calculated based on feature representations across all layers, which demands a finetuning on the whole model during testing. For VOS and LogitNorm, only Direct Test and Simple Adaptive results are reported, as we found it difficult to train them with Domain Adaptation strategy.

\begin{table}[ht]
\centering
\caption{Average results over 10 randomly sampled trajectories on R-MNIST.}\label{table5}
\resizebox{\columnwidth}{!}{\begin{tabular}{@{} c c c c c c @{}}
\toprule
 \multirow{2}{*}{Method} & \multicolumn{2}{c}{R-NOTMNIST} &  \multicolumn{2}{c}{R-Cifar10bw} & \multirow{2}{*}{ID Accuracy} \\
\cmidrule{2-5}
  & \multicolumn{1}{c}{AUROC $\uparrow$} & \multicolumn{1}{c}{FPR95 $\downarrow$} & AUROC $\uparrow$ & FPR95 $\downarrow$  \\
\midrule

Fixed & 97.9 & 13.9 & 98.9 & 9.2  & 35.6 \\
Sampled & 97.7 $\pm$0.4 & 13.6 $\pm$0.8 & 98.8 $\pm$0.4 & 9.6 $\pm$0.7 & 36.7 $\pm$1.4 \\
\bottomrule
\end{tabular}}
\end{table}

\begin{table}[ht]
\begin{center}
\caption{Ablation study on R-MNIST, where w/o $\mathcal{L}_{\rm ood}$ and w/o $\mathcal{L}_{\rm qry}$ are variants trained without $\mathcal{L}_{\rm ood}$ and $\mathcal{L}_{\rm qry}$. $\rm E$ indicates the adding point of $\mathcal{L}_{\rm ood}$, e.g., 1/2 indicates adding $\mathcal{L}_{\rm ood}$ halfway through training.}
\label{table6}
\resizebox{\columnwidth}{!}{\begin{tabular}{ @{}l c c c c  c@{} }
\toprule
 \multirow{2}{*}{Method}  &  \multicolumn{2}{c}{R-NOTMNIST} & \multicolumn{2}{c}{R-Cifar10bw} & \multirow{2}{*}{ID Accuracy} \\
 \cmidrule{2-5}
   & \multicolumn{1}{c}{AUROC $\uparrow$} & \multicolumn{1}{c}{FPR95 $\downarrow$}  & AUROC $\uparrow$ & \multicolumn{1}{c}{FPR95 $\downarrow$} & \\
\midrule
w/o $\mathcal{L}_{\rm ood}$ & 94.1 & 34.9 & 97.6 & 12.2 & 34.2 \\
w/o $\mathcal{L}_{\rm qry}$	& 93.9 & 35.1 &	96.9 & 14.3 & 32.2 \\
E = 1/3 &  99.7 & 7.4	& 99.9 & 5.6 & 28.7 \\
E = 1/2 & 98.0 & 10.4	& 99.1 & 8.7 & 30.1 \\
E = 2/3 &\textbf{96.5}&\textbf{13.9}&\textbf{98.9}&\textbf{9.2}& \textbf{35.6}\\
E = 4/5 &	95.3 &15.5	&95.2&	16.3& 35.9\\
\bottomrule
\end{tabular}}
\end{center}
\end{table}

\noindent \textbf{Evaluation Metrics.} For OOD detection evaluation, we report 1) the area under the receiver operating characteristic curve (AUROC), which measures the model's capacity to distinguish ID and OOD samples based on varying thresholds of the predicted confidence scores; and 2) the false positive rate (FPR95) of OOD samples when the true positive rate of ID samples is fixed at 95\,\%, which measures the percentage of OOD samples that are incorrectly classified as ID samples when the model's sensitivity to ID samples is high.  For evaluating the ID classification, we report the commonly used accuracy.

\subsection{Implementation Details}

On the R-MNIST dataset, we use LeNet as the backbone for all models. As for Cifar10C and Cifar100C datasets, we use WideResNet as the backbone. Note that in our method, the last fully connected layers of LeNet and WideResNet are removed for learning meta-representation during training. We set the uncertainty regularization balancing term $\lambda$ as 0.015 for R-MNIST, and 0.1 for two CifarC datasets. The learning rates in inner/outer loops are set as 0.1/0.01 for R-MNIST, and 0.3/0.03 for two CifarC datasets, respectively. We use SGD with 0.9 momentum and $5 \times 10^{-4}$ weight decay. On each OOD benchmark, all methods are trained using the same backbones with the same number of epochs.

\subsection{Experimental Results}
\noindent \textbf{Results on R-MNIST.} Table \ref{table2} summarizes the ID classification and OOD detection results on R-MNIST. The baseline results show that most existing static OOD detection methods cannot handle ID classification and OOD detection when test samples compass continuous distribution shifts. By using an alternate scheme when we fine-tune the model's classifier during testing (i.e., Simple Adaptive), the comparable methods improved around 2.7\% in ID accuracy and reduced from 5.3\% to 19.3\% in FPR95 across all baselines. Such improvement endorses the dynamic adaptation of the OOD detection model when test samples come from continuously shifting distribution. 

When examining the performance of the Domain Adaptation strategy, results show that ID accuracy improved over the Direct Test baselines, however, the OOD detection performance dropped heavily, from 17.9\% to 26.3\% across all methods in AUROC. In contrast to the Domain Adaptation strategy, our method obtained superior performance (ID accuracy: 35.6\%, AUROC: 97.7\%) by respecting the continuous shifting characteristic of test distributions. Note that although Gram \cite{sastry2020detecting} obtained the best results in FPR95, their ID accuracy (10\% lower than our method) questioned the reliability of the OOD detection performance in CAOOD.

\noindent \textbf{Results on Cifar Corruption.} Table \ref{table3} and \ref{table4} show the OOD detection results on Cifar10C and Cifar100C. Results show that MOL achieved the best performance on both Cifar corruption benchmarks. On Cifar10C, we outperformed Direct Test, Simple Adaptive, and Domain Adaptation strategies by 25.6\%, 7.9\%, and 28.5\% in ID classification respectively. For OOD detection, we outperformed most comparable methods and adaptation strategies. Our AUROC outperformed the second-best result (i.e.,  Simple Adaptive LogitNorm) by 4.2\%. On Cifar100C, we improved the 3 adaptation strategies by 29.6\%, 16.0\%, and 19.7\% respectively in ID accuracy, while maintaining the best average OOD detection performance when compared to all other methods. Note that in our reproduction, Gram shows a catastrophic OOD performance drop and Gradnorm obtained a 100\% in FPR95 when tested on the Cifar100C benchmark, which indicates the instability of their scoring functions using gradients/features across all layers when applied in CAOOD. 

\noindent \textbf{Performance on Continuously Shifting Distribution.} Figure \ref{F3} shows the ID classification and AUROC over distributions on the Rotation MNIST benchmark. It is clear to see that the OOD detection performance of VOS and Energy decreased when the distribution continues to shift, while our method maintained a good OOD detection (i.e., the purple line). This further endorses our method's superiority in addressing the CAOOD problem by learning to quickly adapt to newly arriving test samples with shifting distributions. 

\noindent \textbf{Discussion on the impact of Discretization.}
Theoretically, modeling in continuous time intervals requires infinite samples, making it unrealistic. As such discretization (using finite samples to simulate infinite samples) is a common way to address continuous domain adaptation \cite{wang2022continual, Bobu2018AdaptingTC}. CIDA \cite{wang2020continuously} employs closed intervals which uniformly sample finite time points from these intervals and obtain training data for each sampled time point. Following CIDA, instead of adapting to one fixed discretized trajectory, in meta-training, we adapt to randomly sampled trajectories (i.e., $\mathbf{S}_{\rm spt}$, $\mathbf{S}_{\rm qry}$) to approximate from $0^{\circ}$ to $60^{\circ}$. To further evaluate the impact of discretization on detection accuracy, we evaluate our method on 10 randomly sampled trajectories during testing. Table \ref{table5} shows the average results suggesting that the detection performance remains stable across various trajectories (fixed and sampled).

\subsection{Ablation Study}
A brief ablation analysis is provided in Table \ref{table6}. Firstly, we build a variant of our model without $\mathcal{L}_{\rm ood}$ by taking out the uncertainty regularization term $\mathcal{L}_{\rm ood}$ so it reduces to focus solely on continuous ID adaptation without synthesizing virtual OOD samples for uncertainty regularization. By doing so the ID accuracy slightly dropped by 1.4\% while the OOD detection performance decreases noticeably (3.8\% in AUROC). Then, we conduct experiments on another variant without $\mathcal{L}_{\rm qry}$ without the constraint term on distribution discrepancy $\mathcal{L}_{\rm qry}$, leading to performance drops on all ID accuracy, AUROC, and FPR95. This reflects that harnessing knowledge transfer between distributions is indispensable to the CAOOD problem. Further, we studied the impact on starting point of uncertainty regularization during the whole training process. Results show that an earlier start resulted in a significant drop in ID accuracy while the OOD detection performance is good but meaningless. We suggest that a balance should be identified whenever the ID accuracy is prioritized. More ablations, experimental results, and comparisons about training/testing time refer to the Appendix.

\section{Conclusion and Future Works}
OOD detection has achieved great progress while facing challenges in real-world scenarios when test samples exhibit dynamic distribution shifting. Motivated by these challenges, we propose a novel and more realistic CAOOD detection and develop an effective method of MOL in addressing this problem. To the best of our knowledge, we are the first to investigate OOD detection under continuous distribution shifts. Our proposal will motivate future works to pursue new methods in addressing real-world OOD detection under continuously shifting distributions.

{\small
\bibliographystyle{ieee_fullname}
\bibliography{egbib}
}

\end{document}


\section{Related Works} 
Another related topic is \textbf{Meta-learning} \cite{vilalta2002perspective}, also called \textit{learning-to-learn}, which uses a unique paradigm where a model is trained over a variety of related learning episodes for the benefits of future learning tasks. Such learning-to-learn has been argued to be better aligned with human and animal learning by improving learning on an evolutionary scale \cite{harlow1949formation}. Meta-learning primarily involves learning at two different levels \cite{huisman2021survey}: one of an \textit{inner} level where a new task is presented for quick learning of the training agent, and another of an \textit{outer} level where knowledge has accumulated across earlier inner tasks. Concerned with different learning strategies of inner and outer tasks, Meta-learning approaches can be categorized into three kinds \cite{hospedales2021meta}: optimization-based \cite{finn2017meta}, metric-based \cite{hou2019cross} or model-based \cite{munkhdalai2017meta}, with a wide range of applications on few-short learning \cite{elsken2020meta, hou2019cross}, incremental learning \cite{joseph2021incremental}, reinforcement learning \cite{yu2020meta} and domain generalization \cite{dou2019domain}. 

Model Agnostic Meta-Learning (MAML) \cite{finn2017model} is an optimization-based meta-learning method proposed for fast adaptation of deep neural networks on a small amount of training samples. In OOD detection, one recent work \cite{jeong2020ood} few short \textit{static} OOD detection by directly applying MAML, which is fundamentally different from our proposal in solving CAOOD. Inspired by the idea of learning-to-learn, we link the CAOOD problem to meta-learning by regarding dynamic adaptations over continuously shifting distributions as the inner tasks.

\section{CAOOD Benchmark Details}
Rotation MNIST and Corruption datasets are frequently used in continuous domain adaptation \cite{wang2020continuously,wang2022continual}. In practical applications, rotation can replicate shifts arising from camera positions, while corruptions mimic various weather conditions (e.g., fog, snow) and movements (e.g., motion). In CAOOD, we evaluate both ID accuracy and OOD detection effectiveness. 

\noindent \textbf{Rotation MNIST \cite{deng2012mnist}.} In this benchmark, we use R-MNIST as the ID dataset and use R-Cifar10bw and R-NOTMNIST as the OOD dataset for CAOOD evaluation. Three rotation datasets are derived from their original datasets (i.e., MNIST, cifar10bw, NOTMNIST) by applying the same rotations $0 - 180^{^\circ}$, i.e., $T = \left[0, 180^{\circ}\right]$. MNIST dataset consists of 60000 training samples, and 10000 test samples describing handwritten digits 0-9. NOTMINIST is a near OOD dataset that contains 19000 test samples describing handwritten alphabets A-Z. Cifar10bw is a black-and-white version of Cifar10 with 10000 test samples, which is closer to MNIST. 

The original 60000 labeled ID training samples of size $28 \times 28$ are used as $S$. Then we take the images from rotation $\left(0^\circ, 60^\circ \right]$ as the training shifting samples, i.e., $\{S_{T_k}\}$. In meta-training, we randomly sample $\mathbf{S}_{\rm spt}$ and $\mathbf{S}_{\rm qry}$ of length 10, i.e., $| T_{k}' | = 10$ from $\{S_{T_k}\}$. In meta-testing, for efficient evaluation, we adapt the model sequentially to shifting distributions on rotation ${T_K^-} = \left(120^\circ, 126^\circ, \cdots, 174^\circ \right)$. Note that for each rotation in meta-testing, we have access to only 100 samples. Lastly, we evaluate the ID classification performance on the Rotation MNIST test set, and OOD detection performance on the R-Cifar10bw, and R-NOTMNIST test set. Each test set contains 10000 test samples for every rotation. \\

\noindent \textbf{Cifar10C \cite{hendrycksbenchmarking}} This dataset was initially released for evaluating the robustness of DL models by applying 15 common types of corruptions to Cifar10 \cite{krizhevsky2009learning}, and each type has 5 levels of severities. Concretely it leads to $15 \times 5$ distributions concerned with various corruptions. We use Cifar10C as the ID data and test on two near OOD datasets applying the same shifting distribution: TinyimageNetC, and Cifar100C. Following the OOD benchmark literature \cite{yang2022openood}, we create TinyimageNetC from a subset of the Tinyimagenet \cite{torralba200880} where 1207 images overlap semantic labels with Cifar10 are removed. 

In our protocol, we take clean images from Cifar10 as original labeled training samples $S$, and consider images with various corruptions as shifting samples $\{S^t\}_{T}$. Specifically, we design the continuous shifting distributions by gradually changing the severity across all corruption types, for example: 
$$\footnotesize \underset{\text{t-1 and before}}{\underbrace{{\cdots},\mathrm{C}_{t-1}^{5}}} {\rightarrow}\  \underset{\text{corruption type t, changing gradually}}{\underbrace{\mathrm{C}_{t}^{1},\mathrm{C}_{t}^{2},\mathrm{C}_{t}^{3},\mathrm{C}_{t}^{4},\mathrm{C}_{t}^{5},\mathrm{C}_{t}^{4},\mathrm{C}_{t}^{3},\mathrm{C}_{t}^{2},\mathrm{C}_{t}^{1} }} {\rightarrow}\  \underset{\text{t+1 and on}}{\underbrace{\mathrm{C}_{t+1}^{1},\mathrm{C}_{t+1}^{2},{\cdots}}}$$

\noindent By doing so each corruption type covers 9 shifting distributions with overlaps, resulting in 135 continuously shifting distributions in total, i.e., $| T = 135| $. This reflects a realistic scenario when test samples come from recursive shifting distributions. 

In meta-training, we randomly sample $\mathbf{S}_{\rm spt}$, $\mathbf{S}_{\rm qry}$ of length 10 (i.e., $|T_{k}^{'}|=10$), from the first 7 corruptions of 63 continuously shifting distributions, i.e., $\{S^t\}_{t\in T_k}, |T_k|=63 $. In meta-testing, we adapt our model to a trajectory of length 10 $\left(| T_K^- | = 10\right)$ that is randomly sampled from the last 7 corruptions. Similarly, we only had access to 100 samples in meta-testing. Lastly, we evaluate the ID performance on a trajectory of length 10 from the last 7 corruptions (i.e., frost, fog, brightness, contrast, elastic transform, pixelate, jpeg compression) on the Cifar10C test image, and test the OOD detection performance on TinyimagenetC and Cifar100C test sets. \\

\noindent \textbf{Cifar100C \cite{hendrycksbenchmarking}.} This dataset is similar to Cifar10C by applying the same corruptions to Cifar100. We use Cifar100C as the ID dataset and Cifar10C and TinyimageNetC as the OOD datasets. We re-create TinyimageNetC after 2505 images sharing the same classes with Cifar100 have been removed \cite{yang2022openood}. We applied the same training and testing fashion as used in Cifar10C. 

\section{Virtual OOD Generation Details}
We generate virtual OOD samples in each inner task. Following VOS \cite{duvos}, we maintain an ID class-conditional queue $|Q_y|$ for each class $y \in \mathcal{Y}$ for continuous online estimation of $\hat{\mu}_c^t$ and $\hat{\Sigma}^t$. Specifically, during the starting stage of the training, we en-queue the embeddings of ID features for each class until the queues are filled up. Then, the queues are kept updated by adding new features and deleting the oldest features dynamically. In our experiment, we set $|Q_y|$ to 500 for all three datasets. An effect of $|Q_y|$ on the R-MNIST benchmark is provided in Section D. For the $\delta-\rm likelihood$ region, considering that $\delta$ can be infinitely small, we instead implement by selecting the $p\rm-th$ smallest likelihood in a pool of 1000 samples generated from the estimated Gaussian distribution (per class) \cite{duvos}. Intuitively, a larger $p$ resembles a larger threshold. We set $p=1$ for all experiments.

\section{Detailed Ablation Study}
\noindent \textbf{Effect on Uncertainty Regularization Weight $\lambda$}. In Table \ref{T2} we reported the average OOD detection performance on R-NOTMNIST and R-Cifar10bw when the model was trained on R-MNIST. Generally, a mild uncertainty weight $\lambda $ on $\mathcal{L}^{\rm ood}$ returns an optimal performance, while larger ones over-regularize the model leading to poor ID and OOD performance.

\noindent \textbf{Effect on Queue Size $|Q_k|$}. In Table \ref{T3} we investigate the effect of ID queue size $|Q_y|$ by varying $|Q_y|=\{100,300,500,800\}$. Overall, a larger queue size encourages a more accurate estimation of the ID Gaussian distribution parameters thus benefiting the virtual OOD generation hence improving OOD detection.

\noindent \textbf{Effect on Size of Sampling Pool.} Our method can quickly adapt to newly arriving testing samples for continuously adaptive OOD detection. Note that the size of the sampling pool during testing time could affect the adaptation speed. In Table \ref{T4} we show that a fair-size of sampling pool is enough for fast adaptive OOD detection. A larger size of sampling scale takes more time to finish the adaptation but does not necessarily improve the performance.

\begin{table}[ht]
\begin{center}
\footnotesize
\caption{Ablation study on uncertainty regularization weight $\lambda$. Bold marks the chosen parameter.}
\label{T2}
\begin{tabular}{c | c c c c }
\toprule
$\lambda$ & ID Accuracy$\uparrow$ & AUROC $\uparrow$ & AUPR $\uparrow$& FPR95 $\downarrow$ \\
\toprule
0.01&35.61&	97.50&	98.39&	11.91\\
0.015&	\textbf{35.60}&	\textbf{97.70}&	\textbf{98.42}&	\textbf{11.53}\\
0.02&	33.73&	97.99&	99.01&	10.32\\
0.05&	31.33&	89.70&	87.65&	22.38\\
0.10&	28.51&	47.13&	76.07&	99.99\\
\bottomrule
\end{tabular}
\end{center}
\end{table}

\begin{table}[ht]

\begin{center}
\footnotesize
\caption{Ablation study on the size of queue $| Q_y|$. Bold marks the chosen parameter.}
\label{T3}
\begin{tabular}{c | c c c c }
\toprule
$| Q_y|$ & ID Accuracy$\uparrow$ & AUROC $\uparrow$ & AUPR $\uparrow$& FPR95 $\downarrow$ \\
\toprule
800&	35.60 & 97.98 & 98.01 & 11.31\\
500&	\textbf{35.60}&	\textbf{97.70}&	\textbf{98.42}&	\textbf{11.53}\\
300&	33.97 & 95.66 & 96.62 & 14.85\\
100&	32.67 & 93.42 & 94.63 & 16.17\\
\bottomrule
\end{tabular}
\end{center}
\end{table}

\begin{table}[ht]
\begin{center}
\footnotesize
\caption{Ablation study on the size of sampling pool during the testing process. Bold marks the chosen parameter.}
\label{T4}
\begin{tabular}{c | c c c c | c }
\toprule
Size of Sampling Pool  & ID Accuracy$\uparrow$ & AUROC $\uparrow$ & AUPR $\uparrow$& FPR95 $\downarrow$ & Time Required (seconds) \\ 
\toprule
200& 34.79 & 95.83 & 97.00 & 17.24 & 379.40 \\
500& 35.68 & 95.78 & 95.95 & 15.08 & 422.24 \\
1000& \textbf{35.60 }& \textbf{97.70} & \textbf{98.42} & \textbf{11.53} & 435.87 \\
1500& 35.18 & 96.72 & 97.92 & 12.36 & 500.77 \\ 
5000& 32.60 & 96.04 & 96.27 & 14.19 & 561.94 \\
\bottomrule
\end{tabular}
\end{center}
\end{table}

\section{Comparisons on Training Time} For fair comparisons, we used the same network backbones across all experiments. Table \ref{T5} compares the training time (on 1x A100 GPU). Note that despite a longer training duration, our method achieved significant performance gains, which is particularly crucial for safety-critical scenarios. Such improvements were attained with a brief adaptation period (Table \ref{T4}) in testing, which is practically allowed in terms of gained ID accuracy and OOD detection performance advantages.  

\begin{table}[htbp]
\begin{center}
\footnotesize
\caption{Training time on Energy, VOS, LogitNorm and our method, with ID Accuracy and FPR gaps compared to ours.}\label{T5}
\begin{tabular}{ c | c c c c c c | c }
\toprule
 \multirow{2}{*}{Dataset}  & \multicolumn{2}{c}{Energy} &  \multicolumn{2}{c}{VOS} & \multicolumn{2}{c|}{LogitNorm} & Ours\\
 \cline{2-7}
   & \multicolumn{1}{c}{Time (sec.)} & \multicolumn{1}{c}{Acc. / FPR} &\multicolumn{1}{c}{Time (sec.)} & \multicolumn{1}{c}{Acc. / FPR} & \multicolumn{1}{c}{Time (sec.)} & \multicolumn{1}{c|}{Acc. / FPR} & Time (sec.)\\
\toprule
R-MNIST &	4.7k & -10.0 / -18.3 &	6.8k	& -7.9 / -10.6	& 3.4k	& -10.9 / -16.8	& 15.1k \\
\bottomrule
\end{tabular}
\end{center}
\end{table}

\section{OOD Detection Results on Standard OOD Datasets}
In Table \ref{T6},\ref{T7} and \ref{T8}, we evaluate our methods on commonly used standard OOD datasets from \textit{static} distributions. In this experiment, the test samples consist of continuously shifted ID samples and \textit{static} OOD samples. The results show that existing OOD baselines obtained poor results on standard OOD datasets when the arriving testing ID samples come from continuously shifting distributions. This highlights that distribution shifts on testing ID samples can cause severe damage to OOD detection even if OOD samples are drawn from a static distribution. This may question the generalization ability of current OOD detection methods. 

A simple adaptive adaptation strategy could potentially relieve such impact by adapting the model sequentially to the arriving ID distributions, leading to an improved OOD detection performance. We note that our method outperforms competitive OOD baselines on challenging near-OOD datasets when the OOD samples show similar semantics with the ID datasets.

\begin{table}[htbp]
\begin{center}
\caption{\textbf{OOD detection} on standard OOD datasets (ID dataset: R-MNIST).}
\label{T6}
\resizebox{\textwidth}{!}{\begin{tabular}{c c c c c c c c c c c c c c c c}
\toprule
\multirow{2}{*}{\textbf{Dataset}} & \multicolumn{3}{c}{Energy} &\multicolumn{3}{c}{MSP} &\multicolumn{3}{c}{Simple Adaptive Energy} & \multicolumn{3}{c}{Simple Adaptive MSP} & \multicolumn{3}{c}{\textbf{Ours}}\\
& AUROC	& AUPR &	FPR95&	AUROC&	AUPR&	FPR95&	AUROC&	AUPR&	FPR95&	AUROC&	AUPR&	FPR95&	AUROC	&AUPR&	FPR95\\
\toprule
NOTMNIST&	89.91&	91.05&	45.28&	88.00&	89.60&	50.84&	97.98&	98.13&	25.65&	97.13&	95.07&	20.52&	99.99&	99.99&	0.00\\
Cifar10bw&	97.99&	97.99&	30.33&	99.10&	99.21&	41.93&	99.11&	99.08&	10.29&	99.99&	99.99&	27.09&	100.00&	100.00&	0.00\\
\textbf{Average} &	93.95&	94.52&	37.81&	93.55&	94.41&	46.39&	98.55&	98.61&	17.97&	98.56&	97.53&	23.81&	\textbf{100.00}&	\textbf{100.00}&	\textbf{0.00}\\
\bottomrule
\end{tabular}}
\end{center}
\end{table}

\begin{table}[htbp]
\begin{center}
\caption{\textbf{OOD detection} on standard OOD datasets (ID dataset: Cifar10C).}
\label{T7}
\resizebox{\textwidth}{!}{\begin{tabular}{c c c c c c c c c c c c c c c c}
\toprule
\multirow{2}{*}{\textbf{Dataset}} & \multicolumn{3}{c}{Energy} &\multicolumn{3}{c}{MSP} &\multicolumn{3}{c}{Simple Adaptive Energy} & \multicolumn{3}{c}{Simple Adaptive MSP} & \multicolumn{3}{c}{\textbf{Ours}}\\
& AUROC	& AUPR &	FPR95&	AUROC&	AUPR&	FPR95&	AUROC&	AUPR&	FPR95&	AUROC&	AUPR&	FPR95&	AUROC	&AUPR&	FPR95\\
\toprule
Textures&	43.21&	43.21&	57.24&	45.54&	58.46&	95.06&	64.17&	71.66&	78.77&	65.13&	73.14&	83.98&	69.57&	69.57&	77.13\\
LSUN-C &	71.46&	71.46&	69.43&	66.03&	65.92&	86.69&	87.86&	87.31&	44.98&	81.88&	83.24&	65.62&	85.73&	85.73&	84.44\\
LSUN-R &	54.12&	54.12&	50.99&	54.72&	51.55&	92.01&	87.25&	85.82&	45.25&	81.77&	82.22&	66.82&	87.78&	85.73&	87.13\\
iSUN&	52.69&	52.69&	53.04&	53.87&	53.85&	92.90&	85.50&	84.58&	48.63&	81.08&	83.18&	69.01&	86.47&	86.47&	86.95\\
Places365&	52.43&	52.98&	65.21&	53.08&	54.98&	92.50&	76.98&	77.94&	61.91&	74.80&	76.75&	74.61&	79.08&	78.67&	83.20\\
\textbf{Average Far OOD} &	54.78&	54.89&	59.18&	54.65&	56.95&	91.83&	80.35&	\textbf{81.46}&	\textbf{55.91}&	76.93&	79.71&	72.01&	\textbf{81.73}&	81.23&	83.77\\
\bottomrule
Cifar100&	40.67&	43.43&	95.35&	45.22&	45.13&	95.86&	60.14&	60.34&	91.90&	64.12&	61.98&	87.60&	65.87&	65.87&	80.37\\
TinyImageNet&	59.31&	70.93&	91.18&	60.12&	70.77&	92.32&	68.51&	64.04&	74.40&	68.91&	67.57&	81.45&	67.05&	66.04&	80.01\\
\textbf{Average Near OOD}&	49.99&	57.18&	93.26&	52.67&	57.95&	94.09&	64.32&	62.19&	83.15&	66.52&	64.78&	84.53&	\textbf{66.46}&	\textbf{65.96}&	\textbf{80.19}\\
\bottomrule

\end{tabular}}
\end{center}
\end{table}

\begin{table}[htbp]
\begin{center}

\caption{\textbf{OOD detection} on standard OOD datasets (ID dataset: Cifar100C).}
\label{T8}
\resizebox{\textwidth}{!}{\begin{tabular}{c c c c c c c c c c c c c c c c}
\toprule
\multirow{2}{*}{\textbf{Dataset}} & \multicolumn{3}{c}{Energy} &\multicolumn{3}{c}{MSP} &\multicolumn{3}{c}{Simple Adaptive Energy} & \multicolumn{3}{c}{Simple Adaptive MSP} & \multicolumn{3}{c}{\textbf{Ours}}\\
& AUROC	& AUPR &	FPR95&	AUROC&	AUPR&	FPR95&	AUROC&	AUPR&	FPR95&	AUROC&	AUPR&	FPR95&	AUROC	&AUPR&	FPR95\\
\toprule
Textures&	50.01&	57.17&	94.19&	48.24&	62.65&	93.85&	58.72&	69.72&	87.39&	57.67&	69.82&	91.08&	52.58&	66.51&	91.63\\
LSUN-C&	60.30&	68.18&	96.97&	38.82&	42.19&	99.16&	76.32&	75.91&	69.32&	71.08&	72.07&	79.42&	75.98&	78.53&	77.21\\
LSUN-R&	45.97&	46.75&	89.11&	55.05&	52.19&	90.86&	62.29&	60.22&	80.97&	59.38&	58.53&	87.18&	65.75&	64.53&	84.11\\
iSUN&	52.18&	52.01&	85.07&	57.14&	56.44&	88.36&	59.01&	60.03&	86.71&	56.34&	58.65&	90.19&	64.13&	66.18&	84.48\\
\textbf{Average Far OOD}&	52.12&	56.03&	91.34&	49.81&	53.37&	93.06&	64.09&	66.47&	\textbf{81.10}&	61.12&	64.77&	86.97&	\textbf{64.61}&	\textbf{68.94}&	84.36\\
\toprule
Cifar10	&49.11&	46.31&	98.18&	45.35&	47.03&	97.06&	69.95&	73.24&	78.51&	70.92&	91.93&	84.63&	70.77&	76.85&	73.94\\
TinyImageNet&	47.21&	40.09&	94.97&	42.55&	44.68&	95.95&	70.09&	69.25&	72.02&	66.39&	65.51&	84.09&	71.25&	66.99&	71.75\\
\textbf{Average Near OOD}&	48.16&	43.20&	96.58&	43.95&	45.86&	96.51&	70.02&	71.25&	75.27&	68.66&	\textbf{78.72}&	84.36&	\textbf{71.01}&	71.92&	\textbf{72.85}\\
\bottomrule
\end{tabular}}
\end{center}
\end{table}


\section{Detailed Experimental Results}
\cref{T9,T10,T11,T12,T13,T14,T15,T16,T17} provide detailed ID classification (Accuracy) and OOD detection results (AUROC, AUPR, FPR95) over continuously shifting distributions for each method. 

\begin{table}[htbp]
\footnotesize
\begin{center}
\scriptsize
\caption{\textbf{ID accuracy} on R-MNIST. Post-hoc methods include MSP/Odin/Mahalanobis/Energy/Gram. }
\label{T9}
\begin{tabular}{c c  c  c c c c c c c c c c}
 \toprule
 \textbf{Strategy} & \textbf{Method}& \ $120^{\circ}$ & $126^{\circ}$ & $132^{\circ}$ & $138^{\circ}$ & $144^{\circ}$ & $150^{\circ}$ & $156^{\circ}$ & $162^{\circ}$ & $168^{\circ}$ & $172^{\circ}$ & \textbf{Average} \\
 \toprule
\multirow{2}{*}{Direct Test} &Post-hoc methods & 20.32&	21.66&	22.92&	24.15&	25.28&	26.36&	27.44&	28.38&	29.24&	30.04&	25.58 \\
&VOS& 20.98&	21.99&	22.86&	24.73&	26.98&	29.17&	30.97&	31.51&	32.98&	34.77&	27.69 \\ 
\multirow{2}{*}{Simple Adaptive}&Post-hoc methods  &17.85&	19.86&	22.28&	25.15&	28.04&	29.76&	32.14&	34.66&	36.14&	38.26&	28.41 \\
&VOS &18.47&	21.25&	23.92&	27.23&	29.91&	33.18&	36.03&	38.17&	39.44&	41.38&	30.90\\
\multirow{1}{*}{Domain Adaptation}&Post-hoc methods  &19.44&	19.28&	21.42&	23.16&	25.43&	28.78&	31.09&	33.05&	34.47&	36.31&	27.24 \\
\hline
\multicolumn{2}{c}{\textbf{MOL}}&25.24&	28.28&	31.05&	33.81&	36.28&	38.06&	39.74&	40.38&	41.23&	41.93&	35.60\\
 \bottomrule
\end{tabular}
\end{center}
\end{table}

\begin{table}[htbp]
\footnotesize
\begin{center}
\scriptsize
\caption{\textbf{OOD detection} performance R-NOTMNIST (OOD) when the model was trained on R-MNIST. For each method, we report AUROC$\uparrow$, AUPR $\uparrow$, and FPR95 $\downarrow$ in order.}
\label{T10}
\begin{tabular}{c c  c  c c c c c c c c c c}
 \toprule
 \textbf{Strategy} & \textbf{Method}& \ $120^{\circ}$ & $126^{\circ}$ & $132^{\circ}$ & $138^{\circ}$ & $144^{\circ}$ & $150^{\circ}$ & $156^{\circ}$ & $162^{\circ}$ & $168^{\circ}$ & $172^{\circ}$ & \textbf{Average} \\
 \toprule
\multirow{21}{*}{Direct Test} & \multirow{3}{*}{MSP} & 74.08 	&74.80 &	75.19 	&76.62 &	77.22 &	78.58 & 78.85 	&77.96 	&77.37 	&78.44 &	76.91 \\
 && 76.27 	&7.14 	&77.55 &	78.57 &	79.34 &	80.80 &	80.95 &	79.69 &	78.33 	&79.05 &	78.77 \\
 && 76.64 	&74.91 &	73.45 &	71.85 &	73.58 &	71.62 &	70.66 &	70.46 &	68.53 &	66.96 &	71.87  \\
 \cline{3-13}
 &\multirow{3}{*}{Odin} & 63.07 &	62.90 &	63.26 &63.24 	&63.88 	&65.00 &	64.75 &	64.48 &	62.64 &	62.83 &	63.61  \\
&& 63.26 &	63.51 &	64.26 	&64.45 &	64.62 &	65.20 	&65.13 &	65.19 &	63.55 &	64.02 &	64.32  \\ 
&& 84.32 &	86.78 &	85.49 &	86.06 &	85.79 &	84.79 &	83.74 &	83.82 &	85.42 &	85.40 &	85.16  \\ 
 \cline{3-13}
 &\multirow{3}{*}{Energy} & 90.26&	91.88& 	92.62& 	93.06& 	93.07& 	93.51& 	93.27& 	92.74& 	92.38& 	93.07& 	92.59  \\ 
&&90.52& 	92.30& 	93.11& 	93.48& 	93.54& 	94.07& 	94.00& 	93.50& 	92.96& 	93.51& 	93.10  \\
&&41.87& 	39.42& 	37.17& 	33.30& 	34.26& 	33.74& 	36.22& 	37.06& 	35.75& 	31.90& 	36.07  \\ 
 \cline{3-13}
&\multirow{3}{*}{Mahalanobis} & 79.79& 	85.23& 	85.35& 	89.12& 	88.94& 	90.02& 	91.82& 	92.20& 	91.40& 	92.70& 	88.66 \\
&&84.37& 	89.84& 	89.17& 	91.40& 	91.90& 	91.94& 	92.43& 	92.22& 	90.56& 	91.19& 	90.50 \\
&&85.00& 	71.00& 	65.00& 	62.00& 	71.00& 	60.00& 	36.00& 	28.00& 	40.00& 	23.00& 	54.10 \\
 \cline{3-13}
&\multirow{3}{*}{VOS} &90.37& 	88.73& 	87.89& 	87.69& 	88.30& 	87.04& 	85.72& 	83.26& 	83.30& 	85.15& 	86.74 \\ 
&& 91.06& 	89.46& 	89.01& 	88.94& 	89.66& 	88.63& 	87.36& 	84.99& 	84.67& 	85.91& 	87.97 \\
&& 44.64& 	50.05& 	53.19& 	55.34& 	57.18& 	62.25& 	62.63& 	66.39& 	65.87& 	58.22& 	57.58 \\
 \cline{3-13}
&\multirow{3}{*}{Gram} & 96.14&	95.55&	96.30&	97.09&	95.06&	95.64&	97.50&	96.90&	96.31&	96.08&	96.26\\
&&96.79&	96.10&	95.98&	98.13&	97.67&	96.43&	97.01&	96.33&	97.99&	97.47&	96.99\\
&&5.09&	8.48&	9.22&	11.74&	7.27&	15.59&	12.81&	13.21&	12.68&	9.91&	10.60\\
\hline
\hline
\multirow{21}{*}{Simple Adaptive}&\multirow{3}{*}{MSP} &71.40& 	72.98& 	74.65& 	76.84& 	77.51& 	77.68& 	77.11& 	77.05& 	76.06& 	77.48& 	75.88 \\
&&71.10& 	73.31& 	75.74& 	77.68& 	78.34& 	79.32& 	78.69& 	78.05& 	76.83& 	78.34& 	76.74 \\ 
&&77.62& 	75.69& 	72.95& 	69.56& 	68.09& 	69.33& 	69.09& 	66.82& 	68.00& 	67.17& 	70.43 \\
 \cline{3-13}
&\multirow{3}{*}{Odin} &72.33&	74.13&	77.43&	79.84&	80.90&	79.97&	77.55&	75.90&	74.50&	74.83&	76.74\\
&&72.84&	74.64&	77.82&	80.70&	82.52&	82.08&	79.45&	77.36&	75.36&	75.17&	77.79\\
&&80.09&	77.13&	72.26&	70.18&	67.64&	70.13&	72.14&	72.61&	72.11&	71.60&	72.59\\
 \cline{3-13}
&\multirow{3}{*}{Energy} & 88.75&	90.49&	92.09&	92.70&	93.17&	93.75&	93.13&	92.16&	90.45&	90.76&	91.75\\
&&88.87&	90.96&	92.68&	93.38&	93.88&	94.50&	93.91&	92.84&	90.89&	91.12&	92.30\\
&&49.26&	45.65&	39.87&	37.18&	35.85&	34.80&	37.64&	37.02&	40.60&	39.07&	39.69\\
 \cline{3-13}
&\multirow{3}{*}{Mahalanobis} & 88.75&	90.49&	92.09&	92.70&	93.17&	93.75&	93.13&	92.16&	90.45&	90.76&	91.75\\
&&88.87&	90.96&	92.68&	93.38&	93.88&	94.50&	93.91&	92.84&	90.89&	91.12&	92.30\\
&&49.26&	45.65&	39.87&	37.18&	35.85&	34.80&	37.64&	37.02&	40.60&	39.07&	39.69\\

 \cline{3-13}
&\multirow{3}{*}{VOS}&90.99&	91.82&	93.03&	94.33&	95.36&	96.04&	95.59&	94.26&	92.02&	91.98&	93.54\\
&&90.14&	91.62&	93.27&	94.72&	95.80&	96.52&	96.12&	94.83&	92.57&	92.62&	93.82\\
&&34.00&	36.07&	32.91&	29.76&	25.95&	24.20&	27.61&	33.34&	39.39&	40.05&	32.33\\
\hline
\hline
\multicolumn{2}{c}{\multirow{3}{*}{\textbf{MOL}}}&94.67&	95.37&	96.07&	96.19&	96.62&	96.70&	97.15&	97.21&	97.13&	97.59&	96.47\\
&&96.73&	96.77&	97.38&	97.98&	98.09&	98.21&	98.27&	98.28&	98.24&	98.66&	97.86\\
&&21.65&	19.12&	17.32&	15.05&	12.99&	11.75&	11.24&	9.22&	10.31&	9.83&	13.85\\
 \bottomrule
\end{tabular}
\end{center}
\end{table}

\begin{table}[htbp]
\footnotesize
\begin{center}
\scriptsize
\caption{\textbf{OOD detection} performance on R-Cifar10bw (OOD) when the model was trained on R-MNIST (ID). For each method, we report AUROC$\uparrow$, AUPR $\uparrow$, and FPR95 $\downarrow$ in order.}
\label{T11}

\begin{tabular}{c c  c  c c c c c c c c c c}
 \toprule
 \textbf{Strategy} & \textbf{Method}& \ $120^{\circ}$ & $126^{\circ}$ & $132^{\circ}$ & $138^{\circ}$ & $144^{\circ}$ & $150^{\circ}$ & $156^{\circ}$ & $162^{\circ}$ & $168^{\circ}$ & $172^{\circ}$ & \textbf{Average} \\
 \toprule
\multirow{21}{*}{Direct Test}  &\multirow{3}{*}{MSP} & 94.64&	93.98&	93.03&	91.61&	90.44&	89.28&	88.05&	87.42&	86.38&	86.28&	90.11\\
&&95.67&	95.09&	94.41&	93.36&	92.44&	91.49&	90.46&	89.85&	88.83&	88.44&	92.00\\
&&34.63&	37.22&	44.57&	54.79&	58.27&	61.81&	65.20&	66.77&	67.19&	65.24&	55.57\\
 \cline{3-13}
 &\multirow{3}{*}{Odin} &86.81&	86.74&	86.42&	86.76&	87.02&	88.55&	88.65&	88.23&	87.22&	86.08&	87.25\\
&&88.85&	88.62&	88.34&	88.55&	88.72&	89.97&	90.09&	89.80&	89.02&	87.78&	88.98\\
&&60.52&	59.77&	60.28&	58.33&	58.82&	53.71&	55.33&	56.70&	58.94&	59.81&	58.22\\
 \cline{3-13}
 &\multirow{3}{*}{Energy} &99.11&	99.15&	99.16&	98.95&	98.66&	97.70&	96.76&	95.72&	94.73&	95.10&	97.50\\
&&99.10&	99.13&	99.15&	98.96&	98.70&	97.87&	97.03&	95.89&	94.90&	94.87&	97.56\\
&&4.04&	4.09&	4.04&	4.92&	6.33&	11.98&	18.08&	23.08&	27.12&	22.65&	12.63\\
 \cline{3-13}
&\multirow{3}{*}{Mahalanobis} & 89.75&	89.52&	89.24&	89.45&	89.62&	90.87&	90.99&	90.70&	89.92&	83.23&	89.33\\
&&91.08&	90.86&	90.49&	89.90&	90.31&	91.50&	92.51&	93.22&	93.60&	87.72&	91.12\\
&&60.21&	59.46&	59.97&	58.02&	58.51&	53.40&	55.02&	56.39&	58.63&	61.39&	58.10\\
 \cline{3-13}
&\multirow{3}{*}{VOS}& 97.55&	96.76&	95.88&	94.60&	93.38&	91.40&	90.46&	88.96&	88.76&	89.66&	92.74\\
&&97.51&	96.69&	95.84&	94.51&	93.21&	91.06&	89.98&	88.60&	88.41&	89.15&	92.50\\
&&12.03&	16.42&	20.59&	25.41&	29.13&	35.89&	38.33&	44.61&	46.31&	41.72&	31.04\\
  \cline{3-13}
&\multirow{3}{*}{Gram} & 99.37&	99.38&	99.07&	99.09&	98.42&	98.50&	98.71&	98.84&	98.09&	99.15&	98.86\\
&&99.33&	99.40&	99.13&	98.53&	98.59&	98.85&	99.05&	98.34&	98.11&	99.12&	98.85\\
&&2.91&	3.48&	4.65&	4.27&	7.25&	6.47&	5.22&	3.93&	4.96&	2.86&	4.60\\
\hline
\hline
\multirow{21}{*}{Simple Adaptive}&\multirow{3}{*}{MSP} &87.69&	88.23&	88.67&	88.66&	89.20&	89.98&	90.25&	90.26&	90.20&	90.49&	89.36\\
&&88.45&	89.09&	89.55&	89.56&	90.14&	90.98&	91.25&	91.34&	91.19&	91.55&	90.31\\
&&49.23&	49.66&	48.68&	48.08&	47.80&	48.02&	47.59&	48.78&	48.76&	48.75&	48.54\\
 \cline{3-13}
&\multirow{3}{*}{Odin} &88.52&	87.89&	88.04&	87.69&	87.84&	88.08&	88.09&	88.15&	87.95&	88.29&	88.05\\
&&89.29&	88.77&	88.83&	88.53&	88.57&	88.79&	88.76&	88.67&	88.31&	88.76&	88.73\\
&&48.32&	50.25&	50.10&	51.18&	51.84&	49.75&	51.80&	51.77&	52.08&	52.20&	50.93\\
 \cline{3-13}
&\multirow{3}{*}{Energy} &96.88&	96.66&	96.45&	96.17&	95.95&	96.15&	95.80&	95.67&	94.87&	94.81&	95.94\\
&&96.82&	96.57&	96.31&	95.99&	95.73&	95.95&	95.66&	95.60&	94.85&	94.90&	95.84\\
&&15.79&	16.55&	17.15&	18.04&	19.23&	18.88&	20.58&	21.40&	24.95&	26.26&	19.88\\
 \cline{3-13}
&\multirow{3}{*}{Mahalanobis} & 90.75&	90.50&	90.12&	89.28&	89.55&	90.60&	91.72&	92.33&	92.83&	83.32&	90.10\\
&&92.78&	92.56&	92.19&	91.60&	92.01&	93.20&	94.21&	94.92&	95.30&	89.92&	92.87\\
&&51.62&	52.19&	49.13&	50.71&	48.34&	50.65&	48.12&	50.14&	50.39&	43.71&	49.50\\
 \cline{3-13}
&\multirow{3}{*}{VOS}&97.00&	96.61&	96.29&	96.28&	96.35&	96.94&	97.14&	97.32&	96.84&	96.66&	96.74\\
&&97.01&	96.62&	96.40&	96.43&	96.55&	97.12&	97.29&	97.50&	97.06&	96.90&	96.89\\
&&14.60&	16.41&	17.80&	18.60&	17.92&	15.67&	14.62&	14.23&	16.62&	17.65&	16.41\\
\hline
\hline
\multicolumn{2}{c}{\multirow{3}{*}{\textbf{MOL}}}&99.28&	99.27&	99.21&	99.05&	99.00&	98.92&	98.88&	98.73&	98.53&	98.47&	98.93\\
&&99.27&	99.25&	99.19&	99.03&	98.97&	98.91&	98.87&	98.74&	98.56&	98.50&	98.93\\
&&4.94&	4.67&	4.99&	5.68&	6.40&	7.66&	9.69&	11.85&	16.87&	19.25&	9.20\\
 \bottomrule
\end{tabular}
\end{center}
\end{table}

\begin{table}[htbp]
\footnotesize
\begin{center}
\scriptsize
\caption{\textbf{ID accuracy} on Cifar10C. Post-hoc methods include MSP/Odin/Mahalanobis/Energy/Gram/Gradnorm. The test corruption types cover frost, fog, brightness, contrast, elastic transform, pixelate, and jpeg compression.}
\label{T12}

\begin{tabular}{c c  c  c c c c c c c c c c}
 \toprule
 \textbf{Strategy} & \textbf{Method}& \ ${t_1}$  & ${t_2}$ &${t_3}$ & ${t_4}$ & ${t_5}$ &  ${t_6}$ & ${t_7}$ &  ${t_8}$ & ${t_9}$ & ${t_{10}}$& \textbf{Average} \\
 \toprule
\multirow{2}{*}{Direct Test} &Post-hoc methods &46.85&	41.23&	40.07&	40.96&	37.76&	36.00&	36.40&	34.20&	32.85&	34.84&	38.48 \\
&VOS& 28.42&	27.22&	26.96&	26.77&	26.07&	25.62&	25.26&	24.17&	23.45&	23.82&	25.77 \\ 
\multirow{2}{*}{Simple Adaptive}&Post-hoc methods & 70.38&	53.87&	57.49&	68.44&	42.06&	46.06&	69.24&	32.41&	41.91&	79.67&	56.15 \\
&VOS &37.04&32.51	&31.22&	39.08&	30.51	&30.82&	35.27&	22.85&	25.25&	41.98&	31.62\\
\multirow{1}{*}{Domain Adaptation}&Post-hoc methods &42.81&	33.98&	36.06&	44.08&	28.91&	30.05&	40.84&	23.85&	26.96&	48.62&	35.62 \\
\hline
\multicolumn{2}{c}{\textbf{MOL}}&72.34&	60.74&	63.05&	74.98&	51.42&	55.10&	76.63&	47.29&	66.05&	81.22&	64.18\\
 \bottomrule
\end{tabular}
\end{center}
\end{table}

\begin{table}[htbp]
\footnotesize
\begin{center}
\scriptsize
\caption{\textbf{OOD detection} performance on Cifar100C (OOD) when the model was trained on Cifar10C (ID). For each method, we report AUROC$\uparrow$, AUPR $\uparrow$, and FPR95 $\downarrow$ in order. The test corruption types cover frost, fog, brightness, contrast, elastic transform, pixelate, and jpeg. }
\label{T13}

\begin{tabular}{c c  c  c c c c c c c c c c}
 \toprule
 \textbf{Strategy} & \textbf{Method}& \ ${t_1}$  & ${t_2}$ &${t_3}$ & ${t_4}$ & ${t_5}$ &  ${t_6}$ & ${t_7}$ &  ${t_8}$ & ${t_9}$ & ${t_{10}}$& \textbf{Average} \\
 \toprule
\multirow{21}{*}{Direct Test}  &\multirow{3}{*}{MSP}&57.68&	53.34&	53.94&	55.75&	48.97&	49.93&	56.00&	51.14&	52.24&	60.24&	53.22\\
&&58.48&	53.92&	54.34&	56.61&	49.61&	50.20&	56.21&	50.13&	51.07&	61.50&	53.40\\
&&93.57&	94.32&	94.23&	93.84&	95.83&	95.09&	93.40&	94.10&	93.93&	92.49&	94.26\\

 \cline{3-13}
 &\multirow{3}{*}{Odin} &42.32&	46.66&	46.06&	44.25&	51.03&	50.07&	44.00&	48.86&	47.76&	39.76&	46.08\\
&&45.55&	47.93&	47.79&	46.41&	51.12&	50.12&	46.12&	48.71&	48.10&	43.93&	47.58\\
&&98.00&	97.10&	96.94&	97.82&	95.22&	95.45&	97.39&	94.25&	94.57&	98.71&	96.55\\
 \cline{3-13}
 
 &\multirow{3}{*}{Energy} &57.29&	52.73&	53.65&	55.23&	48.15&	49.09&	55.80&	50.84&	51.89&	60.31&	52.74\\
&&58.16&	53.32&	53.90&	56.08&	49.11&	49.69&	55.77&	49.87&	50.71&	61.25&	52.96\\
&&93.95&	94.61&	94.70&	93.58&	95.64&	95.60&	93.21&	94.27&	94.08&	91.67&	94.40\\
 \cline{3-13}
 
&\multirow{3}{*}{Mahalanobis}& 43.69&	47.01&	46.80&	45.62&	49.65&	49.32&	45.83&	49.15&	48.52&	42.03&	46.76\\
&&46.00&	48.19&	48.21&	47.26&	49.57&	49.36&	46.85&	49.51&	48.75&	44.74&	47.84\\
&&97.69&	97.28&	97.21&	97.92&	95.56&	95.84&	96.88&	95.37&	95.24&	98.32&	96.73\\
 \cline{3-13}
&\multirow{3}{*}{VOS}&53.89&	52.66&	52.60&	54.46&	53.59&	53.90&	52.78&	48.64&	49.80&	54.05&	52.64\\
&&54.84&	53.13&	53.21&	55.64&	54.36&	54.27&	53.66&	49.05&	50.38&	55.11&	53.37\\
&&96.07&	95.92&	95.85&	95.33&	94.88&	94.16&	95.67&	96.73&	96.09&	96.00&	95.67\\
 \cline{3-13}
&\multirow{3}{*}{Gradnorm}&46.32&	13.24&	17.31&	26.47&	6.21&	5.19&	36.31&	32.96&	40.77&	70.42&	29.52\\
&&48.32&	32.92&	33.91&	36.54&	31.63&	31.48&	41.03&	41.82&	100.00&	78.00&	47.56\\
&&97.00&	100.00&	100.00&	98.00&	100.00&	100.00&	100.00&	99.00&	98.68&	79.69&	97.24\\
 \cline{3-13}
&\multirow{3}{*}{Gram}&56.65 & 54.46&54.46 & 55.75&52.66&53.56&57.03&53.43&55.43&58.34&55.18\\
&&50.96&48.88&48.81&49.20&46.54&47.36&50.51&46.62&47.85&71.74&50.85\\
&&88.27&89.93&90.46&88.29&91.83&91.34&88.49&92.54&90.06&86.02&89.72\\
\hline
\hline
\multirow{21}{*}{Simple Adaptive}&\multirow{3}{*}{MSP} &66.71&	60.55&	60.94&	66.97&	57.09&	58.37&	67.63&	54.24&	56.78&	72.63&	62.19\\
&&66.15&	60.33&	60.35&	67.17&	57.91&	59.08&	68.45&	53.97&	57.19&	72.42&	62.30\\
&&89.66&	92.50&	91.49&	90.62&	93.71&	93.58&	90.50&	94.11&	93.48&	86.18&	91.58\\
 \cline{3-13}
&\multirow{3}{*}{Odin}& 66.78&	59.89&	60.75&	66.21&	56.28&	57.08&	66.10&	52.87&	55.30&	71.87&	61.31\\
&&66.30&	59.94&	60.64&	65.93&	56.45&	57.42&	66.40&	52.30&	54.98&	71.32&	61.17\\
&&89.74&	92.73&	92.26&	90.22&	93.83&	93.82&	90.81&	93.93&	94.11&	87.17&	91.86\\
 \cline{3-13}
&\multirow{3}{*}{Energy}&68.07&	60.77&	61.69&	67.22&	56.41&	57.72&	67.29&	52.75&	55.61&	72.93&	60.84\\
&&68.36&	62.34&	62.88&	68.42&	58.62&	58.54&	68.17&	55.98&	58.00&	73.26&	62.37\\
&&88.52&	91.96&	91.35&	89.37&	93.29&	92.87&	89.32&	93.37&	92.82&	84.79&	91.43\\
 \cline{3-13}
&\multirow{3}{*}{Mahalanobis}& 68.45&	62.36&	62.90&	68.31&	57.50&	58.82&	68.24&	55.27&	57.25&	73.53&	63.26\\
&&69.36&	63.81&	63.63&	69.69&	59.13&	60.03&	69.98&	56.06&	58.82&	74.19&	64.47\\
&&89.76&	92.10&	91.32&	89.73&	92.99&	93.39&	89.66&	93.75&	93.35&	86.83&	91.29\\
 \cline{3-13}
&\multirow{3}{*}{VOS}&54.90&	52.41&	52.31&	56.71&	54.09&	54.31&	54.65&	51.05&	51.93&	57.00&	53.60\\
&&57.11&	54.32&	54.06&	59.05&	55.63&	55.66&	56.50&	52.48&	53.36&	59.51&	55.35\\
&&96.47&	96.63&	96.28&	96.24&	95.91&	95.56&	96.91&	96.94&	97.09&	96.63&	96.45\\
\hline
\hline
\multicolumn{2}{c}{\multirow{3}{*}{\textbf{MOL}}}&73.39&	66.73&	68.69&	74.99&	61.67&	62.54&	79.20&	61.36&	65.41&	82.55&	69.65\\
&&73.30&	67.89&	68.28&	75.09&	61.50&	63.30&	77.46&	60.74&	66.06&	81.27&	69.49\\
&&84.49&	88.58&	87.51&	83.08&	91.57&	90.58&	81.27&	91.12&	89.70&	75.48&	86.34\\	
 \bottomrule
\end{tabular}
\end{center}
\end{table}

\begin{table}[htbp]
\begin{center}
\scriptsize
\caption{\textbf{OOD detection} performance on TinyImageNetC (OOD) when the model was trained on Cifar10C (ID). For each method, we report AUROC$\uparrow$, AUPR $\uparrow$, and FPR95 $\downarrow$ in order. The test corruption types cover frost, fog, brightness, contrast, elastic transform, pixelate, and jpeg compression.}
\label{T14}
\begin{tabular}{c c  c  c c c c c c c c c c}
 \toprule
 \textbf{Strategy} & \textbf{Method}& \ ${t_1}$  & ${t_2}$ &${t_3}$ & ${t_4}$ & ${t_5}$ &  ${t_6}$ & ${t_7}$ &  ${t_8}$ & ${t_9}$ & ${t_{10}}$& \textbf{Average} \\
 \toprule
\multirow{21}{*}{Direct Test}  &\multirow{3}{*}{MSP} & 61.50&	60.65&	59.63&	63.68&	60.26&	59.55&	61.66&	54.60&	56.04&	63.65&	59.73\\
&&72.45&	71.08&	70.51&	73.71&	70.93&	70.32&	72.53&	65.78&	66.49&	73.89&	70.42\\
&&92.60&	92.06&	92.35&	91.06&	92.60&	92.82&	91.75&	93.81&	93.10&	91.00&	92.46\\

 \cline{3-13}
 &\multirow{3}{*}{Odin} &38.50&	39.35&	40.37&	36.32&	39.74&	40.45&	38.34&	45.40&	43.96&	36.35&	40.27\\
&&53.24&	53.08&	53.83&	51.53&	53.40&	53.93&	52.84&	56.92&	55.84&	51.70&	53.85\\
&&99.51&	99.26&	99.34&	99.50&	99.28&	99.24&	99.54&	98.24&	98.34&	99.57&	99.14\\
 \cline{3-13}
 
 &\multirow{3}{*}{Energy} &62.86&	61.95&	61.68&	64.87&	62.49&	61.33&	63.56&	56.56&	57.80&	64.59&	61.46\\
&&73.16&	71.57&	71.44&	74.40&	72.19&	71.41&	73.42&	66.87&	67.40&	74.29&	71.32\\
&&91.57&	89.84&	89.41&	90.72&	91.57&	92.37&	91.21&	94.75&	93.59&	90.15&	91.67\\
 \cline{3-13}
 
&\multirow{3}{*}{Mahalanobis}&42.56&	46.93&	47.90&	42.33&	60.17&	58.32&	38.74&	42.51&	43.22&	39.04&	46.97\\
&&55.02&	58.35&	58.67&	54.95&	71.23&	69.13&	52.82&	57.45&	56.32&	52.45&	59.33\\
&&98.79&	98.10&	97.99&	98.59&	96.04&	96.68&	99.13&	99.13&	98.72&	98.94&	98.13\\
 \cline{3-13}
&\multirow{3}{*}{VOS}&57.66&	58.70&	61.18&	58.56&	61.53&	60.93&	56.82&	50.10&	50.57&	54.95&	57.34\\
&&68.35&	68.23&	71.05&	69.16&	72.14&	71.52&	68.66&	62.87&	62.91&	65.71&	68.32\\
&&97.71&	97.28&	97.16&	96.78&	97.03&	96.69&	97.29&	97.41&	97.09&	97.16&	97.16\\
 \cline{3-13}
&\multirow{3}{*}{Gradnorm}&26.41&	27.57&	33.37&	63.24&	61.16&	49.15&	20.43&	21.09&	41.24&	69.79&	41.34\\
&&45.91&	46.12&	50.03&	69.90&	74.68&	63.72&	43.86&	44.71&	32.85&	34.84&	50.66\\
&&100.00&	100.00&	95.59&	86.76&	92.65&	98.53&	100.00&	100.00&	100.00&	91.18&	96.47\\
 \cline{3-13}
&\multirow{3}{*}{Gram}&69.01&59.28&67.12&61.87&89.55&88.75&47.13&46.09&46.73&89.04&66.46\\
&&71.73&62.76&70.09&65.73&92.55&91.75&53.05&51.49&61.59&58.16&67.89 \\
&&79.34&88.12&83.81&86.04&57.95&58.09&93.28&95.33&94.31&89.04&82.53\\

\hline
\hline
\multirow{21}{*}{Simple Adaptive}&\multirow{3}{*}{MSP} &69.42&	64.66&	64.81&	69.57&	59.58&	60.25&	70.65&	57.43&	60.03&	73.71&	64.04\\
&&78.15&	74.25&	74.13&	78.49&	71.25&	71.67&	80.07&	68.83&	71.30&	80.41&	74.24\\
&&89.18&	90.24&	90.31&	90.00&	92.51&	92.47&	89.51&	93.68&	93.25&	85.79&	91.24\\
 \cline{3-13}
&\multirow{3}{*}{Odin}&68.91&	63.93&	64.00&	69.17&	59.03&	59.13&	69.61&	54.67&	58.50&	73.27&	62.99\\
&&77.20&	73.10&	72.97&	77.62&	69.95&	70.12&	78.99&	66.40&	69.51&	80.00&	72.87\\
&&89.29&	90.97&	90.03&	89.59&	92.91&	93.31&	89.84&	93.99&	93.38&	86.96&	91.48\\
 \cline{3-13}
&\multirow{3}{*}{Energy}&70.69&	65.06&	65.68&	70.33&	61.03&	60.86&	72.04&	58.32&	61.18&	74.78&	65.02\\
&&77.85&	73.33&	73.63&	77.90&	70.71&	70.68&	80.22&	68.97&	71.49&	80.27&	73.86\\
&&86.01&	88.56&	87.03&	87.18&	90.00&	90.90&	86.84&	92.10&	92.18&	81.97&	88.98\\
 \cline{3-13}
&\multirow{3}{*}{Mahalanobis}&71.47&	67.02&	67.57&	70.03&	59.72&	60.17&	72.64&	62.56&	62.68&	75.39&	65.98\\
&&80.65&	77.25&	77.42&	79.38&	71.13&	71.52&	82.24&	73.81&	74.09&	82.94&	76.39\\
&&89.12&	90.03&	88.25&	89.03&	92.93&	92.96&	88.21&	92.32&	92.22&	86.06&	90.56\\
 \cline{3-13}
&\multirow{3}{*}{VOS}&61.85&	62.55&	65.85&	60.85&	61.64&	61.06&	59.24&	52.26&	53.77&	59.04&	59.90\\
&&72.32&	72.82&	75.40&	70.99&	70.79&	70.26&	70.51&	66.14&	66.42&	70.07&	70.63\\
&&96.53&	95.84&	94.10&	96.84&	96.75&	96.47&	98.63&	99.53&	98.99&	97.29&	97.08\\
\hline
\hline
\multicolumn{2}{c}{\multirow{3}{*}{\textbf{MOL}}}&76.44&	71.69&	72.82&	75.33&	66.44&	66.07&	79.67&	65.25&	68.89&	81.36&	71.40\\
&&83.88&	80.83&	81.28&	82.75&	74.62&	74.65&	86.65&	75.97&	78.78&	87.19&	79.93\\
&&82.46&	87.49&	85.40&	82.49&	86.53&	86.93&	78.44&	92.18&	88.53&	74.09&	85.60\\ 		
 \bottomrule
\end{tabular}
\end{center}
\end{table}


\begin{table}[htbp]
\footnotesize
\begin{center}
\scriptsize
\caption{\textbf{ID accuracy} on Cifar100C. Post-hoc methods include MSP/Odin/Mahalanobis/Energy/Gram/Gradnorm. The test corruption types cover frost, fog, brightness, contrast, elastic transform, pixelate, and jpeg compression. }
\label{T15}

\begin{tabular}{c c  c  c c c c c c c c c c}
 \toprule
 \textbf{Strategy} & \textbf{Method}& \ ${t_1}$  & ${t_2}$ &${t_3}$ & ${t_4}$ & ${t_5}$ &  ${t_6}$ & ${t_7}$ &  ${t_8}$ & ${t_9}$ & ${t_{10}}$& \textbf{Average} \\
 \toprule
\multirow{2}{*}{Direct Test}&Post-hoc methods &33.69&	28.65&	27.67&	29.55&	27.25&	25.95&	27.19&	25.89&	25.31&	27.22&	27.84\\
&VOS&30.23	&27.59&	26.18&	27.32	&24.08	&24.99	&25.31	&24.18	&24.67&	26.76&26.13\\
\multirow{2}{*}{Simple Adaptive}&Post-hoc methods & 50.24&	36.62&	39.75&	51.21&	27.82&	30.19&	53.94&	27.84&	34.73&	61.79&	41.41\\
&VOS &51.63&	39.34&	42.30&	52.74&	29.71&	31.69&	55.47&	30.81&	36.37&	61.56&	43.16\\
\multirow{1}{*}{Domain Adaptation}&Post-hoc methods &45.41&	33.92&	36.31&	47.71&	27.57&	29.74&	47.20&	23.13&	29.63&	56.34&	37.70 \\
\hline
\multicolumn{2}{c}{\textbf{MOL}}&66.63&	54.34&	57.30&	66.74&	43.71&	45.69&	69.47&	44.81&	50.37&	75.56&	57.46\\
 \bottomrule
\end{tabular}
\end{center}
\end{table}

\begin{table}[htbp]
\begin{center}
\scriptsize
\caption{\textbf{OOD detection} performance on Cifar10C (OOD) when the model was trained on Cifar100C (ID). For each method, we report AUROC$\uparrow$, AUPR $\uparrow$, and FPR95 $\downarrow$ in order. The test corruption types cover frost, fog, brightness, contrast, elastic transform, pixelate, and jpeg compression.}
\label{T16}

\begin{tabular}{c c  c  c c c c c c c c c c}
 \toprule
 \textbf{Strategy} & \textbf{Method}& \ ${t_1}$  & ${t_2}$ &${t_3}$ & ${t_4}$ & ${t_5}$ &  ${t_6}$ & ${t_7}$ &  ${t_8}$ & ${t_9}$ & ${t_{10}}$& \textbf{Average} \\
 \toprule
\multirow{21}{*}{Direct Test}  &\multirow{3}{*}{MSP}& 60.47&	57.11&	57.68&	61.22&	54.62&	55.37&	61.76&	52.90&	55.46&	65.42&	58.20\\
&&62.10&	57.13&	58.49&	62.64&	55.12&	55.92&	63.36&	52.47&	55.24&	67.92&	59.04\\
&&92.69&	92.92&	93.18&	91.93&	93.70&	93.95&	92.22&	94.23&	93.17&	90.51&	92.85\\

 \cline{3-13}
 &\multirow{3}{*}{Odin} &39.53&	42.89&	42.32&	38.78&	45.38&	44.63&	38.24&	47.10&	44.54&	34.58&	41.80\\
&&43.84&	45.38&	45.34&	43.29&	46.66&	46.65&	43.17&	47.93&	46.49&	41.07&	44.98\\
&&98.91&	97.67&	98.19&	98.86&	97.37&	97.74&	99.06&	95.87&	97.00&	99.52&	98.02\\
 \cline{3-13}
 
 &\multirow{3}{*}{Energy}&61.18&	57.88&	58.33&	62.06&	55.02&	55.94&	62.45&	52.65&	55.21&	66.48&	58.72\\
&&62.20&	57.20&	58.58&	62.90&	55.16&	56.10&	63.33&	51.83&	54.61&	68.45&	59.04\\
&&92.60&	91.99&	93.41&	91.94&	93.32&	93.40&	92.66&	94.73&	93.99&	90.89&	92.89\\
 \cline{3-13}
 
&\multirow{3}{*}{Mahalanobis}&38.54&	41.71&	41.41&	37.76&	43.96&	43.51&	36.65&	44.46&	42.26&	33.49&	40.38\\
&&43.87&	45.13&	45.07&	43.12&	46.14&	46.04&	42.49&	46.32&	45.20&	40.93&	44.43\\
&&99.70&	98.36&	98.51&	99.71&	98.23&	98.50&	99.80&	98.33&	98.81&	99.89&	98.98\\
 \cline{3-13}
&\multirow{3}{*}{VOS}&57.66&	58.70&	61.18&	58.56&	61.53&	60.93&	56.82&	50.10&	50.57&	54.95&	57.34\\
&&68.35&	68.23&	71.05&	69.16&	72.14&	71.52&	68.66&	62.87&	62.91&	65.71&	68.32\\
&&97.71&	97.28&	97.16&	96.78&	97.03&	96.69&	97.29&	97.41&	97.09&	97.16&	97.16\\
 \cline{3-13}
&\multirow{3}{*}{Gram}&50.01&49.81&49.85&49.99&49.77&49.66&49.99&48.30&48.09&50.02&49.55\\
&&47.38&47.48&47.53&47.36&47.26&47.31&47.35&46.85&47.19&47.38&47.31\\
&&95.00&95.00&94.99&95.03&95.03&95.32&95.03&97.44&98.17&95.04&95.60\\
\hline
\hline
\multirow{21}{*}{Simple Adaptive}&\multirow{3}{*}{MSP} &66.62&	61.82&	62.26&	66.87&	58.79&	59.02&	69.14&	60.76&	62.87&	72.11&	64.03\\
&&69.25&	64.28&	65.32&	69.49&	59.92&	60.65&	72.11&	62.92&	65.44&	75.01&	66.44\\
&&90.44&	92.38&	92.40&	89.79&	92.13&	92.84&	90.40&	92.34&	91.64&	87.88&	91.22\\
 \cline{3-13}
&\multirow{3}{*}{Odin}&65.85&	60.69&	61.75&	66.27&	57.94&	58.47&	69.33&	59.77&	62.98&	71.94&	63.50\\
&&68.24&	62.83&	64.38&	68.55&	59.03&	59.72&	72.00&	61.10&	64.82&	74.70&	65.54\\
&&89.18&	92.32&	92.24&	89.91&	91.92&	93.13&	88.67&	92.05&	91.17&	87.06&	90.77\\
 \cline{3-13}
&\multirow{3}{*}{Energy}&66.73&	60.80&	61.70&	66.86&	58.42&	59.35&	69.35&	58.60&	61.93&	72.64&	63.64\\
&&69.10&	62.59&	64.27&	69.03&	59.12&	60.22&	71.69&	58.81&	62.93&	75.13&	65.29\\
&&90.95&	92.69&	93.00&	89.90&	92.01&	92.72&	89.28&	93.20&	92.73&	87.12&	91.36\\
 \cline{3-13}

&\multirow{3}{*}{VOS} &57.34&	52.06&	52.61&	57.62&	50.86&	51.49&	59.48&	51.53&	53.95&	61.06&	54.80\\
&&61.33&	56.69&	57.20&	61.62&	55.07&	55.50&	63.44&	55.36&	57.96&	73.93&	59.81\\
&&98.01&	100.00&	100.28&	98.06&	100.43&	99.95&	97.95&	100.24&	99.66&	85.42&	98.00\\
\hline
\hline
\multicolumn{2}{c}{\multirow{3}{*}{\textbf{MOL}}}&67.28&	62.82&	64.63&	66.91&	63.22&	63.42&	66.13&	50.57&	56.04&	69.99&	63.10\\
&&79.65&	75.17&	76.99&	79.12&	75.32&	75.32&	79.68&	67.10&	71.45&	84.09&	76.39\\
&&87.49&	89.24&	88.99&	88.76&	88.12&	87.96&	90.01&	95.66&	93.51&	80.56&	89.03\\
 \bottomrule
\end{tabular}
\end{center}
\end{table}

\begin{table}[htbp]
\begin{center}
\scriptsize
\caption{\textbf{OOD detection} performance on TinyImageNetC (OOD) when the model was trained on Cifar100C (ID). For each method, we report AUROC$\uparrow$, AUPR $\uparrow$, and FPR95 $\downarrow$ in order. The test corruption types cover frost, fog, brightness, contrast, elastic transform, pixelate, and jpeg compression.}
\label{T17}

\begin{tabular}{c c c  c c c c c c c c c c}
 \toprule
 \textbf{Strategy} & \textbf{Method}& \ ${t_1}$  & ${t_2}$ &${t_3}$ & ${t_4}$ & ${t_5}$ &  ${t_6}$ & ${t_7}$ &  ${t_8}$ & ${t_9}$ & ${t_{10}}$& \textbf{Average} \\
 \toprule
\multirow{21}{*}{Direct Test}  &\multirow{3}{*}{MSP}& 55.16&	47.68&	51.04&	56.20&	44.65&	47.26&	54.45&	43.23&	47.67&	61.44&	50.88\\
&&65.07&	57.53&	60.61&	65.47&	55.69&	57.66&	64.53&	55.15&	58.23&	70.83&	61.08\\
&&93.96&	94.65&	94.43&	93.25&	96.01&	95.49&	94.65&	96.81&	95.81&	92.57&	94.76\\

 \cline{3-13}
 &\multirow{3}{*}{Odin} &44.84&	52.32&	48.96&	43.80&	55.35&	52.74&	45.55&	56.77&	52.33&	38.56&	49.12\\
&&56.30&	60.42&	58.66&	55.59&	62.92&	61.13&	57.18&	64.85&	61.55&	52.83&	59.14\\
&&97.81&	93.41&	95.79&	97.44&	92.49&	93.85&	97.49&	93.01&	94.69&	98.72&	95.47\\
 \cline{3-13}
 
 &\multirow{3}{*}{Energy}&56.12&	47.69&	51.49&	55.55&	43.49&	46.40&	54.12&	42.86&	47.33&	61.50&	50.66\\
&&64.99&	57.08&	60.26&	64.61&	54.60&	56.69&	63.75&	54.93&	57.80&	70.46&	60.52\\
&&93.00&	94.65&	94.40&	93.93&	96.07&	95.44&	95.41&	97.21&	96.21&	92.69&	94.90\\
 \cline{3-13}
 
&\multirow{3}{*}{Mahalanobis}&46.37&	50.48&	49.28&	42.74&	57.86&	55.42&	42.56&	54.30&	50.14&	36.19&	48.53\\
&&59.34&	59.82&	60.07&	55.32&	66.16&	64.80&	54.97&	63.25&	59.89&	51.44&	59.51\\
&&98.50&	95.90&	96.74&	98.66&	93.15&	94.59&	99.19&	96.57&	97.79&	99.40&	97.05\\
 \cline{3-13}
&\multirow{3}{*}{VOS}&57.13&	49.07&	50.31&	51.06&	46.97&	43.44&	50.74&	45.77&	48.91&	55.60&	49.90\\
&&64.80&	55.17&	56.93&	59.71&	53.04&	55.58&	65.61&	52.79&	64.03&	69.44&	59.71\\
&&95.11&	94.96&	95.41&	94.90&	98.17&	96.28&	97.01&	98.99&	97.45&	92.02&	96.03\\

 \cline{3-13}
&\multirow{3}{*}{Gram}&54.06&39.78&46.81&49.09&65.09&63.81&37.37&32.52&35.44&47.12&47.11\\
&&60.99&51.22&56.58&56.79&75.19&73.46&48.04&46.93&48.51&54.21&57.19 \\
&&94.47&97.33&95.64&95.38&91.15&92.15&97.92&99.68&99.58&95.80&95.91\\
\hline
\hline
\multirow{21}{*}{Simple Adaptive}&\multirow{3}{*}{MSP}& 65.52&	60.19&	61.22&	66.10&	55.19&	56.59&	67.11&	52.01&	57.62&	71.34&	61.29\\
&&74.70&	69.78&	71.29&	74.92&	65.21&	66.42&	76.46&	63.54&	68.45&	79.38&	71.01\\
&&89.49&	91.69&	91.94&	88.90&	92.57&	92.81&	90.46&	95.37&	93.37&	86.62&	91.32\\
 \cline{3-13}
&\multirow{3}{*}{Odin}&64.89&	57.61&	60.27&	65.25&	52.64&	54.68&	66.76&	48.12&	55.79&	71.12&	59.71\\
&&74.15&	66.78&	69.87&	73.92&	62.86&	64.54&	75.73&	59.27&	65.99&	79.08&	69.22\\
&&89.56&	91.76&	91.26&	89.65&	93.10&	93.29&	89.31&	95.85&	93.71&	86.47&	91.40\\
 \cline{3-13}
&\multirow{3}{*}{Energy}&64.72&	55.19&	58.77&	63.98&	51.56&	54.24&	64.47&	44.23&	52.49&	70.62&	58.03\\
&&73.19&	63.72&	67.65&	72.34&	61.18&	63.46&	73.35&	55.94&	62.62&	78.13&	67.16\\
&&89.44&	92.72&	91.91&	90.15&	92.56&	92.85&	91.47&	97.59&	95.93&	86.74&	92.14\\
 \cline{3-13}
&\multirow{3}{*}{VOS}& 56.81&	50.98&	52.62&	58.09&	48.61&	50.57&	57.01&	39.92&	47.66&	58.72&	52.10\\
&&66.20&	59.18&	61.49&	66.98&	58.33&	59.92&	65.97&	50.74&	56.97&	71.31&	61.71\\
&&95.71&	97.99&	97.03&	94.51&	98.78&	97.63&	96.19&	100.65&	98.99&	85.53&	96.30\\
\hline
\hline

\multicolumn{2}{c}{\multirow{3}{*}{\textbf{MOL}}}&70.28&	65.82&	67.63&	71.91&	68.22&	72.42&	72.13&	63.57&	69.04&	73.77&	69.48\\
&&82.65&	78.17&	79.99&	82.12&	78.32&	78.32&	82.68&	70.10&	74.45&	85.85&	79.27\\
&&86.49&	88.24&	87.99&	87.76&	87.12&	86.96&	89.01&	94.66&	92.51&	86.94&	88.77\\
 \bottomrule
\end{tabular}
\end{center}
\end{table}

\FloatBarrier
{\small
\bibliographystyle{ieee_fullname}
\bibliography{egbib}
}